\documentclass[twoside,11pt]{article}

\usepackage[nohyperref,preprint]{jmlr2e}

\usepackage[utf8]{inputenc} 
\usepackage[T1]{fontenc}    
\usepackage{hyperref}       
\usepackage{url}            
\usepackage{booktabs}       
\usepackage{amsfonts}       
\usepackage{nicefrac}       
\usepackage{microtype}      

\usepackage{savesym,multirow,url,xspace,graphicx,hyperref,enumitem,color}
\usepackage[titletoc]{appendix}
\usepackage{subcaption}
\usepackage{dsfont}
\usepackage{multicol}

\usepackage{etoolbox}

\usepackage{slashbox}

\makeatletter
\newif\if@restonecol
\makeatother

\usepackage[lined, boxed, ruled, commentsnumbered]{algorithm2e}
\usepackage{amsmath,amssymb,xfrac,amsthm}
\setitemize{noitemsep,topsep=1pt,parsep=1pt,partopsep=1pt, leftmargin=12pt}
\newtheorem{lemma_new}{Lemma}
\newtheorem{theorem_new}{Theorem}
\newtheorem*{theorem*}{Theorem}

\newtheorem{corollary_new}{Corollary}
\newtheorem{proposition_new}{Proposition}
\newtheorem{definition_new}{Definition}

\makeatletter

\newcommand{\Rmnum}[1]{\expandafter\@slowromancap\romannumeral #1@}
\makeatother
\usepackage{caption}

\usepackage{algpseudocode}
\usepackage{enumitem}
\usepackage{wasysym}
\usepackage[multiple]{footmisc}

\DeclareMathOperator*{\argmax}{arg\,max}

\newcommand{\costp}{\ensuremath{{C_p}}}
\newcommand{\costr}{\ensuremath{{C_r}}}

\newcommand{\cS}{{S}}

\newcommand{\cA}{\mathcal{A}}

\newcommand{\R}{\mathbb{R}}

\newcommand{\epsP}{\ensuremath{\delta}}
\newcommand{\NoAttack}{\textsc{None}}

\newcommand{\NTJAttack}{\textsc{NT-JAttack}}
\newcommand{\RAttack}{\textsc{RAttack}}
\newcommand{\DAttack}{\textsc{DAttack}}
\newcommand{\JAttack}{\textsc{JAttack}}

\newcommand{\regret}{\textsc{Regret}}
\newcommand{\subopt}{\textsc{SubOpt}}

\newcommand{\cost}{\textsc{Cost}}

\newcommand{\avgmissm}{\textsc{AvgMiss}}
\newcommand{\avgcost}{\textsc{AvgCost}}

\newcommand{\targetpi}{\ensuremath{\pi_{\dagger}}}
\newcommand{\neighbor}[3]{#1\{#2;#3\}}

\newcommand{\stdQ}{\mathcal{Q}}

\usepackage{bm}
\usepackage{bbm}

\newcommand{\norm}[1]{\left\lVert#1\right\rVert}

\newcommand{\expct}[1]{\mathbb{E}\left[#1\right]}

\renewcommand{\Pr}[1]{\ensuremath{\mathbb{P}\left[#1\right] }}

\newcommand{\ind}[1]{\mathds{1}\left[#1\right]}
\newcommand{\pos}[1]{\left[#1\right]^+}

\renewcommand{\cite}{\citep}

\jmlrheading{x}{2020}{x-xx}{11/2020}{xx/xx}{rakhsha20}{Amin Rakhsha, Goran Radanovic, Rati Devidze, Xiaojin Zhu, and Adish Singla}

\ShortHeadings{Policy Teaching in Reinforcement Learning via Environment Poisoning Attacks}{Rakhsha, Radanovic, Devidze, Zhu, and Singla}
\firstpageno{1}

\begin{document}
\title{Policy Teaching in Reinforcement Learning\\via Environment Poisoning Attacks\thanks{~~This manuscript is an extended version of the paper \citep{rakhsha2020policy} that appeared in ICML'20.}}

\author{\name Amin Rakhsha \email arakhsha@mpi-sws.org \\
       \addr Max Planck Institute for Software Systems (MPI-SWS)\\
       Saarbrucken, 66123, Germany\\
       \AND
       \name Goran Radanovic \email gradanovic@mpi-sws.org \\
       \addr Max Planck Institute for Software Systems (MPI-SWS)\\
       Saarbrucken, 66123, Germany\\
       \AND
       \name Rati Devidze \email rdevidze@mpi-sws.org \\
       \addr Max Planck Institute for Software Systems (MPI-SWS)\\
       Saarbrucken, 66123, Germany\\
       \AND
       \name Xiaojin Zhu  \email jerryzhu@cs.wisc.edu \\
       \addr University of Wisconsin-Madison\\
       Madison, WI 53706, USA\\
       \AND
       \name Adish Singla \email adishs@mpi-sws.org \\
       \addr Max Planck Institute for Software Systems (MPI-SWS)\\
       Saarbrucken, 66123, Germany\\      
       }
\editor{x}

\maketitle

\newtoggle{longversion}
\settoggle{longversion}{true}

\begin{abstract}
We study a security threat to reinforcement learning where an attacker poisons the learning environment to force the agent into executing a target policy chosen by the attacker. As a victim, we consider RL agents whose objective is to find a policy that maximizes reward in infinite-horizon problem settings. The attacker can manipulate the rewards and the transition dynamics in the learning environment at training-time, and is interested in doing so in a stealthy manner. We propose an optimization framework for finding an optimal stealthy attack for different measures of attack cost. We provide lower/upper bounds on the attack cost, and instantiate our attacks in two settings: (i) an offline setting where the agent is doing planning in the poisoned environment, and (ii) an online setting where the agent is learning a policy with poisoned feedback. Our results show that the attacker can easily succeed in teaching any target policy to the victim under mild conditions and highlight a significant security threat to reinforcement learning agents in practice.
\end{abstract}


\section{Introduction}\label{sec.introduction}
Understanding adversarial attacks on learning algorithms is critical to finding security threats against the deployed machine learning systems and in designing novel algorithms robust to  those threats. We focus on \emph{training-time} adversarial attacks on learning algorithms, also known as data poisoning attacks \cite{huang2011adversarial,biggio2018wild,zhu2018optimal}. Different from \emph{test-time} attacks where the adversary perturbs test data to change the algorithm's decisions,  poisoning attacks manipulate the training data to change the algorithm's decision-making policy itself.

Most of the existing work on data poisoning attacks has focused on supervised learning algorithms~\cite{DBLP:conf/icml/BiggioNL12,mei2015using,DBLP:conf/icml/XiaoBBFER15,alfeld2016data,li2016data,DBLP:journals/corr/abs-1811-00741}. 
In contemporary works, researchers have explored data poisoning attacks against stochastic multi-armed bandits \cite{DBLP:conf/nips/Jun0MZ18,DBLP:conf/icml/LiuS19a} and contextual bandits \cite{DBLP:conf/gamesec/MaJ0018}, which belong to family of online learning algorithms with limited feedback---such algorithms are widely used in real-world applications such as news article recommendation~\cite{DBLP:conf/www/LiCLS10} and web advertisements ranking~\cite{DBLP:journals/tist/ChapelleMR14}. The feedback loop in online learning makes these applications easily susceptible to data poisoning, e.g., attacks in the form of  click baits \cite{miller2011s}. 

In this paper, we focus on data poisoning attacks against reinforcement learning (RL) algorithms, an online learning paradigm for sequential decision-making under uncertainty~\cite{sutton2018reinforcement}.\footnote{Poisoning attacks is also mathematically equivalent to the formulation of machine teaching with teacher being the adversary~\cite{zhu2018overview}. However, the problem of designing optimized environments for teaching a target policy to an RL agent is not well-understood in machine teaching.}
 Given that RL algorithms are increasingly used in critical applications, including cyber-physical systems~\cite{li2019reinforcement} and personal assistive devices \cite{rybski2007interactive}, it is of utmost importance to investigate the security threat to RL algorithms against different forms of poisoning attacks.


%
%
%

\subsection{Overview of our Results and Contributions}\label{sec.introduction.contributions}
In the following, we discuss a few of the types/dimensions of poisoning attacks in RL in order to  highlight the novelty of our work in comparison to existing work.\footnote{This paper extends the earlier version of the conference paper \citep{rakhsha2020policy} in the following ways. \textbf{(i)} The earlier version studied attacks by poisoning either rewards only or transitions only. In this paper, we introduce a general optimization framework for jointly poisoning the rewards and transitions. We provide new theoretical analysis for the joint attack and empirically show that it leads to more cost-effective attack strategies. \textbf{(ii)} The earlier version studied attacks only against RL agents who maximize average reward in undiscounted infinite-horizon settings. In this paper, we generalize these results by additionally considering discounted infinite-horizon settings. This generalization in turn makes our attack strategies applicable to a broader family of RL agents (e.g., an agent using the Q-learning algorithm). \textbf{(iii)} We provide a detailed discussion on the efficiency of solving the attack optimization problems.}

\paragraph{Type of adversarial manipulation.} Existing works on poisoning attacks against RL have studied the manipulation of rewards only  \cite{DBLP:conf/aaai/ZhangP08,DBLP:conf/sigecom/ZhangPC09,ma2019policy,DBLP:conf/gamesec/HuangZ19a,xuezhou2020adaptive}.
However, for certain applications, it is more natural to manipulate the transition dynamics instead of the rewards, such as (i) the inventory management problem setting where state transitions are controlled by demand and supply of products in a market and (ii) conversational agents where the state is represented by the history of conversations.
%
A key novelty of our work is that we study environment poisoning, i.e., jointly manipulating rewards and transition dynamics. We propose a general optimization framework for environment poisoning; our theoretical analysis provides technical conditions which ensures attacker's success and gives lower/upper bounds on the attack cost.

\paragraph{Objective of the learning agent.} Existing works have focused primarily on studying RL agents that maximizes \emph{cumulative reward in discounted} infinite-horizon settings. However, for many real-world applications, it is more appropriate to consider RL agents that maximizes \emph{average reward in undiscounted} infinite-horizon settings~\cite{Puterman1994,DBLP:journals/ml/Mahadevan96}---in particular, this is a more suitable objective for applications that have cyclic tasks or tasks without absorbing states, e.g., inventory management and scheduling problems \cite{tadepalli1994h,Puterman1994}, and a robot learning to avoid obstacles \cite{DBLP:journals/ai/MahadevanC92}. 
In our work, the proposed optimization framework and theoretical results cover both the optimality criteria--- average reward criteria and discounted reward criteria---in infinite-horizon settings.

\paragraph{Offline planning and online learning.} Most of the existing works have focused on attacks in an \emph{offline} setting where the adversary first poisons the reward function in the environment and then the RL agent finds a policy via \emph{planning} \cite{DBLP:conf/aaai/ZhangP08,DBLP:conf/sigecom/ZhangPC09,ma2019policy}. In contrast, we call a setting as \emph{online} where the adversary interacts with a \emph{learning} agent to manipulate the feedback signals. One of the key differences in these two settings is in measuring attacker's cost: The $\ell_\infty$\emph{-norm} of manipulation is commonly studied for the offline setting; for the online setting, the cumulative cost of attack over time (e.g., measured by $\ell_1$\emph{-norm} of manipulations) is more relevant and has not been studied in literature. A recent contemporary work \cite{xuezhou2020adaptive} considers the online setting, however, does not study $\ell_1$\emph{-norm} of manipulations as the attack cost.
%
We instantiate our attacks in both the offline and online settings with appropriate notions of attack cost. 
\looseness-1We note that our attacks are constructive, and we provide numerical simulations to support our theoretical statements. Our results demonstrate that the attacker can easily succeed in teaching (forcing) the victim to execute the desired target policy at a minimal cost.
\subsection{Additional Related Work} \label{sec.relatedwork}

\paragraph{Test-time attacks against RL.} 
A growing body of contemporary works have studied test-time attacks against RL~\cite{chen2019adversarial}, in particular, on RL algorithms with neural network policies \cite{mnih2015human,schulman2015trust}. These attacks are typically done by adding noise in the observed state (e.g., a camera image) to fool the neural network policy into taking malicious actions~\cite{huang2017adversarial,DBLP:conf/ijcai/LinHLSLS17,tretschk2018sequential}. Different attack goals have been considered in these works, e.g., guiding the agent to some adversarial states or forcing agent to take actions that maximizes adversary's own rewards. 
Our work is technically quite different and is focused on training-time attacks where the goal is to force the agent to learn a target policy.


\paragraph{Teaching an RL agent.} 
\looseness-1Poisoning attacks is mathematically equivalent to the formulation of machine teaching with teacher being the adversary~\cite{goldman1995complexity,singla2013actively,singla2014near,zhu2015machine,zhu2018overview,DBLP:conf/nips/ChenSAPY18,mansouri2019preference,DBLP:conf/nips/PeltolaCDK19,DBLP:conf/ijcai/DevidzeMH0S20}. In particular, there have been a number of recent works on teaching an RL agent via providing an optimized curriculum of demonstrations \cite{cakmak2012algorithmic,DBLP:conf/uai/WalshG12,hadfield2016cooperative,DBLP:conf/nips/HaugTS18,DBLP:conf/ijcai/KamalarubanDCS19,DBLP:conf/nips/TschiatschekGHD19,brown2019machine}. However, these works have focused on imitation-learning based RL agents who learn from provided demonstrations without any reward feedback \cite{osa2018algorithmic}. Given that we consider RL agents who find policies based on rewards, our work is technically very different from theirs. There is also a related literature on changing the behavior of an RL agent via \emph{reward shaping} \cite{ng1999policy,asmuth2008potential}; here the reward function is changed to only speed up the convergence of the learning algorithm while ensuring that the optimal policy in the modified environment is unchanged.
\section{Environment and RL Agent}\label{sec.preliminaries}
We consider a standard RL setting, based on Markov decision processes and RL agents that optimize their expected utility. In a unified manner, we will cover two cases in which RL agents are optimizing their total discounted rewards or their undiscounted average reward.
The following subsections will introduce our setting in more detail.
	
\subsection{Environment, Policy, and Optimality Criteria}\label{sec.preliminaries.env}
The environment is a Markov Decision Process (MDP) defined as $M = (S, A, R, P, \gamma)$, where $S$ is the state space, $A$ is the action space, $R\colon S\times A \to \R$ is the reward function, $P\colon S\times A\times S \to [0, 1]$ is the state transition dynamics, i.e., $P(s, a, s')$ denotes the probability of reaching state $s'$ when taking action $a$ in state $s$, and $\gamma \in [0, 1]$ is the discounting factor, and $d_0$ is the initail state distribution. Note that discount factor $\gamma$ can be equal to $1$, which we treat as a special case, as explained in the paragraphs below. 

We consider a class of deterministic policies: we denote a generic deterministic policy by $\pi$, and we define it as a mapping from states to actions, i.e., $\pi\colon S\to A$. Furthermore, we assume that MDP $M$ is {\em ergodic}, which implies that every policy $\pi$ has a {\em state} distribution $\mu^\pi$ defined as:
\begin{align}
\mu^\pi(s) = 
\begin{cases}
(1 - \gamma) \sum_{t = 0}^{\infty} \gamma^t\Pr{s_t = s | s_0 \sim d_0, \pi} &\quad \mbox{if} \; \gamma < 1\\
\lim_{N\to \infty} \frac{1}{N} \sum_{t=0}^{N-1} \Pr{s_t = s | s_0 \sim d_0, \pi}
&\quad \mbox{if} \; \gamma = 1,
\end{cases}
\end{align}
which satisfies $\mu^\pi(s) > 0$ for every state $s$~\cite{Puterman1994}. For $\gamma = 1$, $\mu^\pi$ corresponds to the {\em stationary} state distribution induced by policy $\pi$, whereas for $\gamma < 1$, $\mu^\pi$ corresponds to the {\em discounted} state distribution induced by policy $\pi$. State distribution $\mu^\pi$ satisfies the following Bellman flow constraints: 
\begin{align}\label{eq.stat_dist.from.kernel}
\mu^{\pi}(s) = (1 - \gamma) \cdot d_0(s) + \gamma \cdot \sum_{s'} P(s', \pi(s'), s) \cdot \mu^{\pi}(s').
\end{align}

Given an initial state distribution $d_0$, the expected average and discounted reward of policy $\pi$ are respectively equal to
$$\lim_{N\to \infty} \frac{1}{N} \expct{\sum_{t=0}^{N-1} R(s_t, a_t)|s_0 \sim d_0, \pi} \textnormal{ and } \expct{\sum_{t=0}^{\infty} \gamma^t R(s_t, a_t)|s_0 \sim d_0, \pi},$$ where the expectations are taken over the rewards received by the agent when starting from initial state $s_0 \sim d_0$ and following the policy $\pi$. 
In our work, we cover both the \emph{average reward}~\cite{Puterman1994,DBLP:journals/ml/Mahadevan96} and  the \emph{discounted reward}  optimality criteria in infinite-horizon settings~\cite{Puterman1994,sutton2018reinforcement}. Throughout the paper, the special case of $\gamma = 1$ represents the average reward problem setting, whereas $\gamma < 1$ represents the discounted reward problem setting.
We unify these cases using a single \emph{score} of policy $\pi$ defined as
\begin{align}\label{eq.score}
    \rho(\pi, M, d_0) := \sum_{s}\mu^\pi(s)\cdot R(s, \pi(s)).
\end{align}
Using policy scores $\rho$, we define the notion of optimality used in this paper. 
A policy $\pi^*$ is optimal if for every other deterministic policy $\pi$ we have $\rho^{\pi^*} \ge \rho^{\pi},$ and $\epsilon$-robust optimal if $\rho^{\pi^*} \ge \rho^{\pi} + \epsilon$ also holds.
%
As shown in prior work (and discussed later in the paper), score $\rho$ is closely related to the standard notions of state-action and state value functions, i.e., $Q$-values and $V$-values. For policy $\pi$, (shifted) $Q$-values and $V$-values are defined as\footnote{
To facilitate exposure of our results in a unified manner, in the discounted reward setting, we are using $Q$-values that are shifted from the standard definition by $\rho^\pi/(1 - \gamma)$. This modification allows using the same Bellman equations for both the average and discounted rewards settings. When referring to standard $Q$-values, we use the symbol $\stdQ$ instead (e.g., as used in the appendices).
}
\begin{align*}
	Q^\pi(s, a) = \expct{\sum_{t = 0}^\infty\gamma^t \cdot (r_t - \rho^\pi) | s_0 = s, a_0 = a, \pi}, \quad V^\pi(s) = Q^\pi(s, \pi(s)),
\end{align*}
and they satisfy the following Bellman equations:
 \begin{gather}
 \label{equation.bellman}
     Q^\pi(s, a) = R(s, a) - \rho^\pi + \gamma \cdot \sum_{s' \in S}P(s, a, s') \cdot V^\pi(s').
 \end{gather}

Finally, we introduce quantities that measure {\em connectedness} of MDP $M$. First, we define a coefficient
$
\alpha = \min_{s,a,s',a'} \sum_{x \in S}\min \big (P(s,a,x), P(s',a',x) \big )
$, so that $(1-\alpha)$ is equivalent to Hajnal measure of $P$~\citep{Puterman1994}.\footnote{As discussed in \citep{Puterman1994}, the Hajnal measure of a Markov chain transition matrix provides an upper bound on its subradius (the modulus of the second largest eigenvalue). Hence, this measure is informative about the mixing times in MDP $M$.}  
Second, we define the notion of {\em discounted reach times} for policy $\pi$ as $T^\pi(s, s) = 0$ and $T^\pi(s, s') = \expct{\sum_{i = 0}^{L^\pi(s, s') - 1}\gamma^{i}}$
for $s' \ne s$, where $L^\pi(s, s')$ is a random variable that counts the number of steps it takes to visit state $s'$ for the first time starting from $s$ and following $\pi$ in $M$. The maximum discounted reach time for policy $\pi$ is denoted by $D^{\pi} = \max_{s,s'} T^{\pi}(s,s')$, i.e., $D^{\pi}$ denotes the diameter of Markov chain induced by policy $\pi$ in MDP $M$. Note that reach times $T^\pi$ satisfy the following recursive equations for $s \ne s'$:
\begin{align*}
T^{\pi}(s, s') &= \sum_{s'' \in S} P(s, \pi(s), s'') \cdot (1 + \gamma \cdot  T^{\pi}(s'', s')),
\end{align*}
which means that for a given MDP $M$ and policy $\pi$, one can compute them efficiently since this is just a system of linear equations. 



%

\subsection{RL Agent}\label{subsec.online_agent}
\begin{figure*}[t!]
\centering
	\begin{subfigure}[b]{0.65\textwidth}
	   \centering
		\includegraphics[width=1\linewidth]{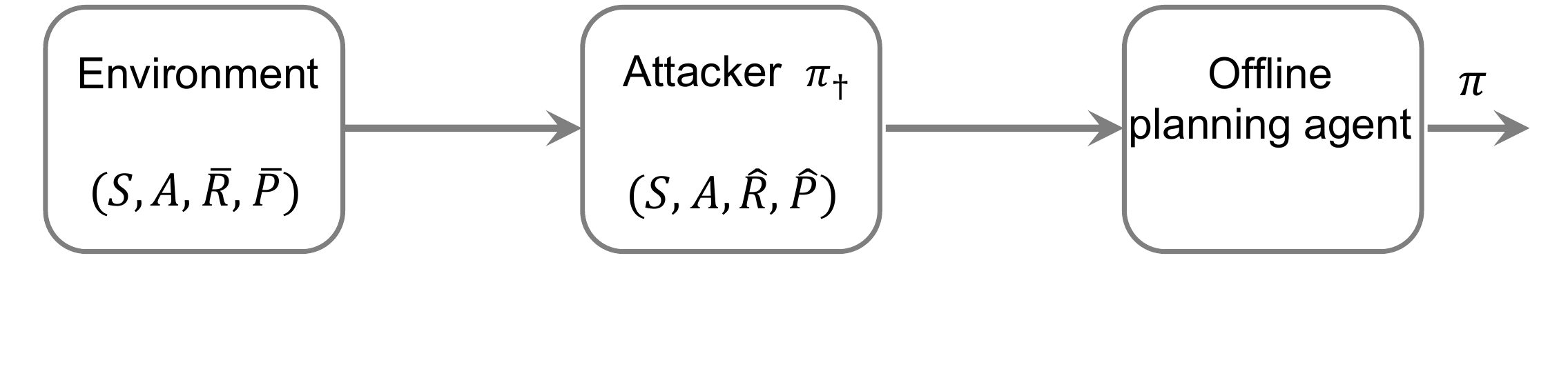}
		\vspace{-10mm}
		\caption{Poisoning attack against an RL agent doing offline planning}
		\vspace{4mm}
		\label{fig:model.offline}
	\end{subfigure}
	\begin{subfigure}[b]{0.65\textwidth}
	    \centering
		\includegraphics[width=1\linewidth]{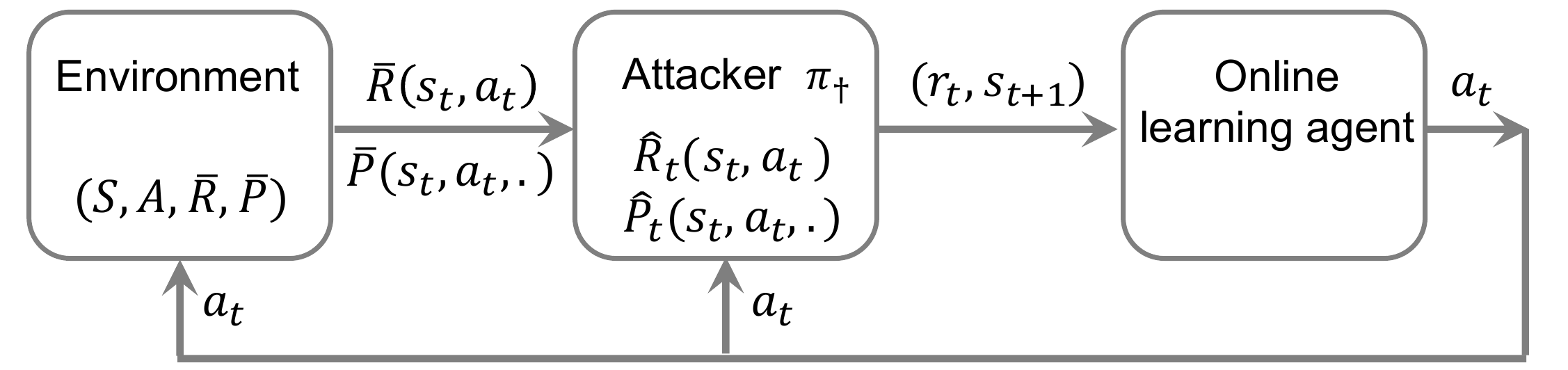}
		\vspace{-4mm}
		\caption{Poisoning attack against an RL agent doing online learning}
		\vspace{2mm}
		\label{fig:model.online}
	\end{subfigure}
   \caption{(a) Adversary first poisons the environment by manipulating reward function and transition dynamics, then, the RL agent finds an optimal policy via \emph{planning} algorithms based on Dynamic Programming \cite{Puterman1994,sutton2018reinforcement}. (b) Adversary interacts with an RL agent to manipulate the feedback signals; here, we consider an agent who is learning a policy based on feedback received from the environment (see Section~\ref{subsec.online_agent} for details).}
	\label{fig:model}
\end{figure*}

We consider RL agents in the following two settings (also, see Figure~\ref{fig:model}).

\looseness-1
\paragraph{Offline planning agent.}
In the \emph{offline} setting, an RL agent is given an MDP $M$, and chooses a deterministic optimal policy $\pi^* \in \argmax_{\pi} \rho(\pi, M, d_0)$. 
The optimal policy can be found via \emph{planning} algorithms based on  Dynamic Programming such as value iteration~\cite{Puterman1994,sutton2018reinforcement}. 

\paragraph{Online learning agent.}
In the \emph{online} setting, an RL agent does not know the MDP $M$ (i.e., $R$ and $P$ are unknown). At each step $t$, the agent stochastically chooses an action $a_t$ based on the previous observations, and then as feedback it obtains reward $r_t$ and transitions to the next state $s_{t+1}$. 
In this paper, we consider agents with two performance measures:
\begin{enumerate}
    \item 
For the case of average reward criteria with $\gamma = 1$, we consider a regret-minimization learner. Performance of a regret-minimization learner in MDP $M$ is measured by its \emph{regret} which after $T$ steps is given by
 $\regret(T, M) = \rho^* \cdot T - \sum_{t=0}^{T-1} r_t$,
where $\rho^* := \rho(\pi^*, M)$ is the optimal score.
Well-studied algorithms with sublinear regret exist for average reward criteria, e.g., UCRL algorithm~\cite{auer2007logarithmic,jaksch2010near} and algorithms based on posterior sampling method~\cite{agrawal2017optimistic}. For more details, we refer the reader to Appendix~\ref{appendix_background}.

\item 
For the case of discounted reward criteria with $\gamma < 1$, the type of learners we consider are evaluated based on the number of suboptimal steps they take.
An agent is suboptimal at time step $t$ if it takes an action not used by any near-optimal policy. This is formulated as   $\subopt(T, M, \epsilon') = \sum_{t=0}^{T-1} \ind{a_t \notin \{ \pi(s_t)~|~\rho^\pi \ge \rho^{\pi^*} - \epsilon' \} }$ where $\ind{.}$ denotes the indicator function and $\epsilon'$ measures near-optimality of a policy w.r.t. score $\rho$.
Our analysis of attacks is based on $\expct{\subopt(T, M, \epsilon')}$ of the learner for a specific value of $\epsilon'$. Some bounds on this quantity are known for existing algorithms such as classic Q-learning~\cite{even2003learning} and Delayed Q-learning~\cite{strehl2006pac} as discussed in more detail in Appendix~\ref{appendix_background}.

\end{enumerate}

\section{Attack Models and Problem Formulation}\label{sec.formulation}
\looseness-1In this section, we formulate the problem of adversarial attacks on the RL agent in both the offline and online settings. In what follows, the original MDP (before poisoning) is denoted by $\overline{M} = (S, A, \overline{R}, \overline{P}, \gamma)$, and an \emph{overline} is added to the corresponding quantities before poisoning, such as $\overline{\rho}^\pi$, $\overline{\mu}^\pi$, and $\overline{\alpha}$.

\looseness-1
The attacker has a target policy $\targetpi$ and poisons the environment with the \emph{goal} of teaching/forcing the RL agent to executing this policy.\footnote{Our results can be translated to attacks against the Batch RL agent studied by \cite{ma2019policy} where the attacker poisons the training data used by the agent to learn MDP parameters.}
The attacker is interested in doing a stealthy attack with minimal \emph{cost} to avoid being detected.
We assume that the attacker knows the original MDP $\overline{M}$, i.e., the original reward function $\overline{R}$ and state transition dynamics $\overline{P}$. This assumption is standard in the existing literature on poisoning attacks against RL. 
The attacker requires that the RL agent behaves as specified in Section~\ref{sec.preliminaries}, however the attacker does not know the agent's algorithm or internal parameters.

\subsection{Attack Against an Offline Planning Agent}\label{sec.formulation.offline}
In attacks against an offline planning agent, the attacker manipulates the original MDP $\overline{M} = (S, A, \overline{R}, \overline{P}, \gamma)$ to a poisoned MDP  $\widehat{M} = (S, A, \widehat{R}, \widehat{P}, \gamma)$ which is then used by the RL agent for finding the optimal policy, see Figure~\ref{fig:model.offline}.

\paragraph{Goal of the attack.} Given a margin parameter $\epsilon$, the attacker poisons the reward function and the transition dynamics so that the target policy $\targetpi$ is $\epsilon$-robust optimal in the poisoned MDP $\widehat{M}$, i.e., the following condition holds:
\begin{align}\label{prob.off.gen_constraints}
    \rho(\targetpi, \widehat{M}, d_0) \ge \rho(\pi, \widehat{M}, d_0) + \epsilon, \quad \forall \pi\ne\pi_\dagger.
\end{align}
\looseness-1

\paragraph{Cost of the attack.} 
To define the cost of an attack, we quantify for each state-action pair how much the attack changes the reward and transition dynamics associated to this state-action pair, i.e., $|\widehat{R}(s,a) - \overline{R}(s,a)\big|$ and $\sum_{s'}\big|\widehat{P}(s,a,s') - \overline{P}(s,a,s')\big|$ respectively. The cost for that state action-pair is the weighted sum of these values, with weights given by two parameters,  $\costr$ and  $\costp$ respectively. The total cost is then measured as the $\ell_p$-norm (for $p \ge 1$) of the costs across all state-action pairs. Formally, this can be written as
%
\begin{align*}
\cost(\widehat{M}, \overline{M}, \costr, \costp, p) = \bigg(\sum_{s,a} \Big(
\costr \cdot \big|\widehat{R}(s,a) - \overline{R}(s,a)\big|
+ 
\costp \cdot \sum_{s'}\big|\widehat{P}(s,a,s') - \overline{P}(s,a,s')\big|
\Big)^p
\bigg)^{1/p}.
\end{align*}
We will write the cost as $\cost(\widehat{M}, \overline{M})$ when other parameters are clear from the context. By taking the limits of $Cp$ (resp. $Cr$) to infinity, the cost function enforces the attacker to poison only the rewards  (resp. only the transitions).

\subsection{Attack Against an Online Learning Agent}\label{sec.formulation.online}
In attacks against an online learning agent, the attacker at time $t$ manipulates the reward function $\overline{R}(s_t, a_t)$ and transition dynamics $\overline{P}(s_t, a_t, .)$ for the current state $s_t$ and the agent's action $a_t$, see Figure~\ref{fig:model.online}. Then, at time $t$,  the (poisoned) reward $r_t$ is obtained from $\widehat{R}_t (s_t,a_t)$ instead of $\overline{R}(s_t,a_t)$ and the (poisoned) next state $s_{t+1}$ is sampled from $\widehat{P}_t(s_t,a_t,.)$ instead of $\overline{P}(s_t,a_t,.)$.

\paragraph{Goal of the attack.} Specification of the attacker's goal in this online setting is not as straightforward as that in the offline setting, primarily because the agent might never converge to any stationary policy. In our work, at time $t$ when the current state is $s_t$, we measure the mismatch of agent's action $a_t$ w.r.t. the target policy $\targetpi$ as $\ind{a_t \ne \targetpi(s_t)}$. With this, we define a notion of \emph{average mismatch} of learner's actions in time horizon $T$ as follows:
\begin{align}\label{eq.avgmis}
    \avgmissm(T) = \frac{1}{T} \cdot \bigg(\sum_{t=0}^{T-1} \ind{a_t \ne \targetpi(s_t)}\bigg).
\end{align}
\looseness-1
The goal of the attacker is to  minimize $\avgmissm(T)$.

\paragraph{Cost of the attack.} 
We consider a notion of \emph{average cost} of attack in time horizon $T$ denoted as $\avgcost(T)$. This is defined as
\begin{align*}
\frac{1}{T} \cdot \bigg(\sum_{t=0}^{T-1} \Big(
\costr \cdot \big|\widehat{R}_t(s_t,a_t) - \overline{R}(s_t,a_t)\big|
+ 
\costp \cdot \sum_{s'}\big|\widehat{P}_t(s_t,a_t,s') - \overline{P}(s_t,a_t,s')\big|
\Big)^p\bigg)^{1/p}, 
\end{align*}
where the $\ell_p$-norm (for $p \ge 1$) is defined over a vector of length $T$ with values quantifying the attack cost at each time step $t$. One of the key differences in measuring attacker's cost for offline and online settings is the use of appropriate norm. While the $\ell_\infty$\emph{-norm} of manipulation is more suitable and commonly studied for the offline setting; for the online setting, the cumulative cost of attack over time measured by the $\ell_1$\emph{-norm} is more relevant.

\section{Attacks in Offline Setting}\label{sec.offlineattacks}
\label{sec.off.ideas-definitions}

	In this section, we introduce and analyze attacks against an offline planing agent that derives its policy using a poisoned MDP $\widehat M$.  The attacker tries to minimally change the original MDP $\overline{M}$, while at the same time ensuring that the target policy is optimal in the modified MDP $\widehat M$.

\subsection{Offline Attacks: Key Ideas and Attack Problem}\label{sec.off.problem}

The main obstacle in performing offline attacks as formulated in Section \ref{sec.formulation.offline} is the complexity of the optimization problem it leads to. To find MDP $\widehat M$ for which the target policy is $\epsilon$-robust optimal, one could directly utilize constraints expressed by \eqref{prob.off.gen_constraints}. However, the number of constraints in \eqref{prob.off.gen_constraints} equals $(|A|^{|S|}-1)$, i.e., it is exponential in $|S|$, making optimization problems that directly utilize them intractable. We show that it is enough to satisfy these constraints for $(|S|\cdot|A|-|S|)$ policies that we call \textit{neighbors} of the target policy and which we define as follows:
\begin{definition_new}
	For a policy $\pi$, its \textit{neighbor} policy $\neighbor{\pi}{s}{a}$ is defined as
	$$
	\neighbor{\pi}{s}{a}(x) = \left\{\begin{array}{lc}
		\pi(x) & x \ne s \\
		a & x = s
	\end{array}\right. .
	$$
\end{definition_new}
The following lemma provides a simple verification criterion for examining whether a policy of interest is $\epsilon$-robust optimal in a given MDP. Its proof can be found in Appendix~\ref{appendix.off.general}.
\begin{lemma_new}
	\label{lemma.using_neighbors}
	Policy $\pi$ is $\epsilon$-robust optimal \emph{iff} we have $\rho^{\pi} \ge \rho^{\neighbor{\pi}{s}{a}} + \epsilon$ for every state $s$ and action $a \ne \pi(s)$.
\end{lemma_new}
In other words, Lemma \ref{lemma.using_neighbors} implies that (sub)optimality of the target policy can be deduced by examining its neighbor policies.  
Using the definition of score $\rho$, Bellman flow constraints (i.e.,  equation \eqref{eq.stat_dist.from.kernel}), and Lemma~\ref{lemma.using_neighbors}, 
we can formulate the problem of modifying MDP $\overline{M} = (S, A, \overline{R}, \overline{P})$ to MDP $\widehat M = (S, A, \widehat{R}, \widehat{P})$ as the following optimization problem:
\begin{align}
\label{prob.off}
\tag{P1}
&\quad \min_{M, R, P,  \mu^{\targetpi}, \mu^{\neighbor{\targetpi}{s}{a}}} \quad \cost(M, \overline{M}, \costr, \costp, p) \\
\notag
&\quad \mbox{ s.t. } \quad \text{$ \mu^{\targetpi}$ and $P$ satisfy \eqref{eq.stat_dist.from.kernel}}, \\
\notag
&\qquad \quad \quad \forall s, a \ne \targetpi(s): \text{$ \mu^{\neighbor{\targetpi}{s}{a}}$ and $P$ satisfy \eqref{eq.stat_dist.from.kernel}}, \\
\notag
&\qquad \quad \quad \forall s, a \ne \targetpi(s): 
\sum_{s'} \mu^{\targetpi}(s') \cdot R \big (s', \targetpi(s') \big ) \ge
\sum_{s'} \mu^{\neighbor{\targetpi}{s}{a}}(s') \cdot  R \big (s', \neighbor{\targetpi}{s}{a}(s') \big ) + \epsilon,\\
\notag
&\qquad \quad \quad \forall s, a, s': P(s,a,s') \ge \epsP \cdot \overline P(s,a,s'),\\
\notag
&\qquad \quad \quad M = (S, A, R, P, \gamma).
\end{align}
Here, $\epsP \in (0, 1]$ in the last set of constraints is a given parameter, specifying how much one is allowed to decrease the original values of transition probabilities. $\epsP > 0$ is a regularity condition which ensures that the new MDP is ergodic.\footnote{\looseness-1This follows because strictly positive trajectory probabilities in MDP $\overline M$ remain strictly positive in the new MDP $\widehat M$, which further implies that all states remain recurrent.} 
In Appendix~\ref{appendix.off.joint}, we provide a more detailed discussion on $\epsP$ and how to choose it.  In the next section, we analyze this problem in detail, providing bounds and discussions on its solution.

%

\subsection{Offline Attacks: Theoretical Analysis}\label{sec.off.analysis}

 We start our analysis by defining quantities relevant for stating our formal results. 
Notice that we denote the quantities defined w.r.t. to the original MDP $\overline{M}$ by putting an {\em overline}. For example, $\overline{V}^\pi(s)$ and $\overline{Q}^\pi(s, a)$ denote $V$-values and $Q$-values of policy $\pi$ in MDP $\overline{M}$. As a measure of the relative optimality gap between the target policy $\targetpi$ and its neighbor policies $\neighbor{\targetpi}{s}{a}$,
we define the following state-action dependent variable:
\begin{align}
    \label{chi.definition}
    \overline \chi^{\pi}_{\epsilon}(s, a) = 
    \begin{cases}
        \pos{\frac{
                \overline \rho^{\neighbor{\pi}{s}{a}} - \overline \rho^{\pi} + \epsilon
            }{
                \overline \mu^{\neighbor{\pi}{s}{a}}(s)
            }} \quad & \textnormal{ for } a\ne\pi(s), \\
            0 &  \textnormal{ for }  a = \pi(s).
    \end{cases}
\end{align}
Here, $\pos{x}$ is equal to $\max\{0, x\}$. As we show in our formal results, $\chi^{\pi}_{\epsilon}$ with an appropriately set $\epsilon$ captures how much one should change state-action values $\overline{Q}^{\targetpi}(s, a)$ relative to state values $\overline{V}^{\targetpi}(s)$ in order to obtain a successful attack.
 %
%
To simplify our formal statements, we also set 
\begin{align}
 \overline \beta(s, a) =
 \epsilon
 \cdot
 \overline \mu^{\neighbor{\targetpi}{s}{a}}(s)
 \cdot
 \frac
 {1 + \gamma \cdot \overline{D}^{\targetpi}}
 {1 - (1 - \gamma) \cdot \overline{D}^{\targetpi}}.
 \end{align}
 Furthermore, we utilize the span of value function, defined as $sp(\overline V^{\targetpi}) = \max_s \overline V^{\targetpi}(s) - \min_s \overline V^{\targetpi}(s)$. In order to define the other quantities of interest, we order states by their values. In particular, $s_i$  denotes an order of states for $1\le i \le |S|$ such that $\overline V^{\targetpi}(s_i)$ is decreasing with $i$. This order allows us to define two important state-action dependent quantities $\overline F_i$ and $\overline G_i$ as
 \begin{align*}
\overline F_i(s, a) &= \gamma \cdot \sum_{j = 1}^{i} (1 - \delta)\cdot \overline P(s, a, s_j)(\overline V^{\targetpi}(s_j) - 
\overline V^{\targetpi}(s_{|S|})), \\
\overline G_i(s, a) &= 2\cdot\sum_{j = 1}^{i} (1 - \delta)\cdot\overline P(s, a, s_j).
\end{align*}
for $a \ne \targetpi(s)$, and $\overline F_i(s, \targetpi(s)) = \overline G_i(s, \targetpi(s)) = 0$ otherwise. We further set $F_0(s,a) = 0$ and $G_0(s,a) = 0$.
Finally, for each state-action pair, we define a number $k(s, a)$ to be the largest element in $\{1, ..., |S|\}$  such that  $\gamma \cdot \costr \cdot (\overline V^{\targetpi}(s_{k(s, a)}) - \overline V^{\targetpi}(s_{|S|})) > 2\cdot\costp$ and $\overline F_{k(s, a)}(s, a) \le \overline{\chi}^{\targetpi}_{\overline\beta(s,a)}(s, a)$. If these conditions cannot be satisfied, we set $k(s, a) = 0$ (in our results, this case would correspond to changing only rewards). We provide a more detailed discussion on these quantities later in the section.  
We can now state the main result for the offline attack setting.

\begin{theorem_new}
	\label{theorem.off.joint}
	If $\widehat{M}$ is an optimal solution to optimization problem (\ref{prob.off}), then 
	\begin{align*}
	\frac{1 - \gamma + \gamma \cdot \delta \cdot \overline{\alpha}}
	{2\cdot\costr^{-1} + \gamma \cdot \costp^{-1} \cdot sp(\overline V^{\targetpi})}
	\norm{\overline{\chi}^{\targetpi}_0}_\infty \le
	\cost(\widehat{M}, \overline{M}) \le
	\norm{\costp \cdot \overline{G}_k + \costr \cdot (\overline{\chi}^{\targetpi}_{\overline \beta} - \overline F_k)}_p,
	\end{align*}{}
	where $\overline{\chi}^{\targetpi}_{\overline \beta}$, $\overline G_k$, and $\overline F_k,$ are  vectors of length $|S|\cdot |A|$ with components $\overline{\chi}^{\targetpi}_{\overline\beta(s, a)}(s, a)$,  $\overline G_{k(s, a)}(s, a)$, and $\overline F_{k(s, a)}(s, a)$.
\end{theorem_new}{}
This theorem gives lower and upper bounds on the cost of offline attacks against a planning agent.  By taking the limits of $\costp$ (resp. $\costr$) to infinity, we can obtain the bounds for the attack which only changes rewards (resp. transitions). As discussed in the earlier version of the paper \cite{rakhsha2020policy}, we note the following two points: (i) the attack that only poisons transitions might not always be feasible, (ii) one can obtain a slightly tighter lower bound for the attack that only poisons rewards. In the following proof sketch, we provide some intuition on how the bounds Theorem \ref{theorem.off.joint} are derived. 

\begin{proof}[Proof Sketch of Theorem~\ref{theorem.off.joint}]

We split the proof-sketch in two parts, corresponding to the upper and the lower bound respectively. 

\paragraph{Upper bound on the cost. }
We provide a constructive upper bound by introducing an attack which is a solution to optimization problem \eqref{prob.off}. The main idea behind this approach is that at each state $s$, we can make each action $a \ne \targetpi(s)$ less valuable than action $\targetpi(s)$ by decreasing $\overline{ R}(s, a)$ (rewards) and changing the next state distribution (transitions). The attack that we consider first modifies the transition dynamics, until the point when it becomes less cost effective to change transitions than the rewards. In the second phase, the attack changes only the rewards. 

To make an action less valuable, the attacker can decrease the probability of transitioning to a high-value state and increase the probability of transitioning to a low-value state, with state values being obtained from value function $\overline V^{\targetpi}(s)$. To construct an upper bound, we analyze an attack that for each state-action pair $(s, a)$ decreases the probability of transitioning to the top $k(s, a)$ highest valued states and increases probability $\overline P(s, a, s_{|S|})$ by the amount that is equal to the total decrease. Note that this attack should not violate the ergodicity constraint, i.e., $\overline P(s, a, s_i)$ can be decreased by at most $(1-\delta) \cdot \overline P(s, a, s_i)$. For the considered attack, the transition probabilities that correspond to the $k(s, a)$ highest valued states are maximally decreased, i.e., we modify $\overline P(s, a, s_i)$ to $\delta \cdot \overline P(s, a, s_i)$ for $i \le k(s, a)$. We now argue that $k(s, a)$ as specified in the theorem accounts for two important factors: minimizing the amount of change and optimizing the efficiency of changes.

First, note that we should not change the original MDP more than it is needed. When changing transitions of state-action pair $(s, a)$ for $k(s, a)$ highest-valued states using the approach described above, the state-action value function of that pair, i.e., $\overline{Q}^{\targetpi}(s, a)$, decreases by $\overline F_{k(s, a)}(s, a)$. As we show in our analysis, 
 $\overline{\chi}^{\targetpi}_{\overline \beta(s,a)}(s, a)$ captures how much state-action values, i.e., 
 $\overline{Q}^{\targetpi}(s, a)$, should be decreased in order for the attack to be successful. This means that we should select $k(s,a)$ so that $\overline F_{k(s, a)}(s, a)$ does not exceed $\overline{\chi}^{\targetpi}_{\overline\beta(s,a)}(s, a)$, i.e., $\overline F_{k(s, a)}(s, a) \le \overline{\chi}^{\targetpi}_{\overline\beta(s,a)}(s, a)$.

 Second, note that we need to account for the efficiency of the modifications. Let us first consider the case of $k(s, a)  > 0$. The difference $V^{\targetpi}(s_{k(s, a)}) - V^{\targetpi}(s_{|S|})$ should be large enough so that modifying transitions $\overline P(s, a, \cdot )$ is more efficient than modifying reward $\overline R(s, a)$.
 Since state-action values $\overline{Q}^{\targetpi}(s, a)$ need to be decreased by $\overline{\chi}^{\targetpi}_{\overline \beta(s,a)}(s, a)$, the notion of {\em efficiency} in this context expresses how much $\overline{Q}^{\targetpi}(s, a)$ changes
  per the unit cost of changing $\overline P(s, a, .)$ and $\overline R(s, a)$ respectively. 
  From equation \eqref{equation.bellman}, we can see that the attack efficiency of changing reward $R(s, a)$ is $\frac{1}{C_r}$, whereas the attack efficiency of changing transition $\overline P(s, a, s_k)$ is $\gamma \cdot \frac{1}{2 \cdot C_p} \cdot (\overline V^{\targetpi}(s_{k(s, a)}) - \overline V^{\targetpi}(s_{|S|}))$.
  Combining this with the definition of the cost of the attack gives us that $k(s, a )$ should satisfy $\gamma \cdot \costr \cdot (\overline V^{\targetpi}(s_{k(s, a)}) - \overline V^{\targetpi}(s_{|S|})) > 2\cdot\costp$.
 %
 If this condition is not possible to satisfy, then $k(s,a)$ is equal to $0$, which corresponds to the attack that changes only rewards. 
 
 To summarize, we pick the largest $k(s, a)$ such that  $\costr \cdot (\overline V^{\targetpi}(s_{k(s, a)}) - \overline V^{\targetpi}(s_{|S|})) > 2\cdot\costp$ and $\overline F_{k(s, a)}(s, a) \le \overline{\chi}^{\targetpi}_{\overline\beta(s,a)}(s, a)$. In case these conditions are infeasible, we chose $k(s, a) = 0$. 
 We then maximally decreases the probability of transitioning to the $k(s, a)$ highest-valued states when taking action $a$ in state $s$, and accordingly increases the probability of transitioning to state $s_{|S|}$. This attack on transitions modifies $\overline{P}(s, a, .)$ by 
 $\overline G_{k(s,a)}(s, a)$, which in turn incurs the cost of $\costp \cdot \overline{G}_{k(s, a)}(s, a)$ and decreases $\overline{Q}^{\targetpi}(s, a)$ by $\overline F_{k(s,a)}(s, a)$.
 To obtain the desired decrease in $Q$-values (i.e., $\overline{\chi}^{\targetpi}_{\overline \beta(s,a)}(s, a)$), we decrease $\overline{R}(s, a)$ by $\overline{\chi}^{\targetpi}_{\overline \beta(s,a)}(s, a) - \overline F_{k(s, a)}(s, a))$, incurring the cost of $ \costr \cdot (\overline{\chi}^{\targetpi}_{\overline \beta(s, a) }(s, a)- \overline F_{k(s, a)}(s, a))$. This gives us an upper bound on the cost of an optimal solution to optimization problem \ref{prob.off}.

\paragraph{Lower bound on the cost.} Similar to our upper bound, our lower bound on the cost mainly depends on three aspects of the problem: the efficiency of rewards poisoning, the efficiency of transitions poisoning, and the difference in state-action values between the target policy $\targetpi$ and other policies. The attack efficiency of poisoning rewards depends only on $\costr$. However, the attack efficiency of poisoning transitions depends both on $\costp$ and the discrepancy among the state values, which are bounded by $sp(\overline V^{\targetpi})$.
Intuitively, if the attack efficiency is low (resp. high), the lower bound on the cost needed to make an attack successful will be high (resp. low). Furthermore, the lower bound depends on the difference in state-action values between the target policy $\targetpi$ and its neighbor policies, as captured by $\norm{\overline{\chi}^{\targetpi}_0}_\infty$. 
Notice that while the upper bound is based on a specific attack, the lower bound is attack-agnostic and implies that any successful attack must incur this cost.
\end{proof}

The full proof can be found in Appendix~\ref{appendix.off.joint}. 

\subsection{Offline Attacks: Efficiency of 
Solving the Problem}\label{sec.off.efficiency}

In the previous subsections, we formulated the problem of attacking an offline RL agent as optimization problem \eqref{prob.off}.
 This problem is difficult to solve in general due to the first three constraints, which are non-linear and render the problem non-convex.   
 However, note that 
 in prior work on these attacks, the special case where only rewards are poisoned is shown to be a convex optimization problem in both the average reward and discounted reward optimality criteria \citep{rakhsha2020policy,ma2019policy}.
 \begin{proposition_new}
 	\label{preposition.off.ronly}
 	\citep{rakhsha2020policy,ma2019policy}
 	The special case of offline attacks in which only rewards can be poisoned by the attacker, i.e. $\widehat P = \overline P$, is solvable through a convex optimization problem.
 \end{proposition_new}
 
  As we discuss in Section~\ref{sec.on.problem.refromulation}, the problem \eqref{prob.off} also becomes convex if two additional constraints are added: $\widehat{R}(s, \targetpi(s)) = \overline{R}(s, \targetpi(s))$ and $\widehat{P}(s, \targetpi(s), .) = \overline{P}(s, \targetpi(s), .)$. Notice that these two constraints restrict the form of a solution in that the corresponding attack is not allowed to manipulate rewards and transition for state-action pairs $(s, \targetpi(s))$. We refer to such attacks as {\em non-target only}, and we will discuss them further in Section~\ref{sec.onlineattacks}.
\section{Attacks in Online Setting}\label{sec.onlineattacks}

\looseness-1We now turn to attacks on an agent that learns over time using the environment feedback. 
Unlike the planning agent from the previous section, an online learning agent derives its policy from the interaction history, i.e., tuples of the form  $(s_{t}, a_{t}, r_{t}, s_{t+1})$. To attack an online learning agent, an attacker changes the environment feedback, i.e.,  reward $r_t$ and state $s_{t+1}$. 

\subsection{Online Attacks: Key Ideas and Attack Problem}\label{sec.on.problem}
 The underlying idea behind our approach is to utilize the fact that the policies of the learning agents that we consider (see Section~\ref{subsec.online_agent}) will converge towards an optimal policy, and therefore will take a bounded number of suboptimal actions. Hence, to steer a learning agent towards selecting the target policy, it suffices to replace the environment feedback (i.e., reward $r_t$ and the next state $s_{t+1}$) with a feedback sampled from an MDP that has the target policy as its $\epsilon$-robust optimal policy. Notice that such an MDP can be obtained using optimization problem \eqref{prob.off}.
Now we separately consider the two cases: i) average reward criteria with $\gamma = 1$ and ii) discounted reward criteria with $\gamma < 1$.

For the case of average reward criteria with $\gamma = 1$, we consider a regret-minimization learner. With the following lemma we show that the above approach is sound: 
assuming that a learner draws its experience from an ergodic MDP $M$ that has $\targetpi$ as its $\epsilon$-robust optimal policy, the expected number of steps in which the learner deviates from $\targetpi$ is bounded by $O(\expct{\regret(T, M)})$. 
\begin{lemma_new}
	\label{lemma.on.known.nontarget}
	(Lemma~2 in \cite{rakhsha2020policy})
	Consider an ergodic MDP $M$ with $\gamma = 1$ that has $\targetpi$ as its $\epsilon$-robust optimal policy, and an online learning agent whose expected regret in MDP $M$ is $\expct{\regret(T, M)}$. The average mismatch of the agent w.r.t. the policy $\targetpi$  is bounded by
	\begin{align}\label{eq_def_K}
	\expct{\avgmissm(T)} \le  \frac{\mu_{\textnormal{max}}}{\epsilon \cdot T} \cdot \Big(\expct{\regret(T, M)} + 2\norm{V^{\targetpi}}_\infty\Big),
	\end{align}
	with $\mu_{\textnormal{max}} := \max_{s,a}\mu^{\neighbor{\targetpi}{s}{a}} (s)$. Here, $\mu^{\pi}$ and $V^{\pi}$ are respectively the stationary distribution and the value function of  policy $\pi$ in MDP $M$.
\end{lemma_new}


For the case of discounted reward criteria with $\gamma < 1$, we consider a learner with bounded number of suboptimal steps. The following lemma is an analog to Lemma \ref{lemma.on.known.nontarget} and is proven in Appendix~\ref{appendix.on}. This lemma is based on the simple observation: when a learner draws its experience from an MDP $M$ that has $\targetpi$ as its $\epsilon$-robust optimal policy, instantiating $\subopt(T, M, \epsilon')$ with $\epsilon' = \epsilon$ will give us the number of times the learner deviates from $\targetpi$. 
\begin{lemma_new}
	\label{lemma.on.subopt.missmatch}
	Consider an ergodic MDP $M$ with $\gamma < 1$ that has $\targetpi$ as its $\epsilon$-robust optimal policy, and an online learning agent whose expected number of suboptimal steps in an MDP $M$ is $\subopt(T, M, \epsilon')$. The average mismatch of the agent w.r.t. the policy $\targetpi$ is given by $\avgmissm(T) = \frac{1}{T} \cdot \subopt(T, M, \epsilon)$.
\end{lemma_new}


To conclude, if a learner has sublinear  $\expct{\regret(T, M)}$ (resp. $\expct{\subopt(T, M, \epsilon)}$), Lemma~\ref{lemma.on.known.nontarget} (resp.  Lemma~\ref{lemma.on.subopt.missmatch}) implies that $o(1)$ average mismatch can be achieved in expectation using a sampling based attack that replaces the environment feedback (sampled from the original MDP $\overline{M}$) with a poisoned feedback sampled from MDP $\widehat M$, where MDP $\widehat M$ is a solution to optimization problem \eqref{prob.off}.


However, the expected average cost of such an attack could be $\Omega(1)$ (non-diminishing over time), even for a learner with sublinear 
$\expct{\regret(T, M)}$
or
$\expct{\subopt(T, M, \epsilon)}$.
Intuitively, if a learner follows the target policy and there exists $s$ for which $\widehat R(s, \targetpi(s)) \ne \overline{R}(s, \targetpi(s)) $  or $\widehat{P}(s, \targetpi(s), .) \ne \overline{P}(s, \targetpi(s), .)$, then the attacker would incur a non-zero cost whenever the learner visits $s$.  
To avoid this issue, we need to enforce constraints on the sampling MDP $\widehat M$ specifying that the attack does not alter rewards and transitions that correspond to the state-action pairs of the target policy, i.e., $(s, \targetpi(s))$. As mentioned in Section~\ref{sec.off.efficiency}, we refer to such attacks as {\em non-target only}.
%
This brings us to the following template that we utilize for attacks on an online learner:
\begin{itemize}
    \item Modify the optimization problem \eqref{prob.off} by adding constraints $\widehat R(s, \targetpi(s)) = \overline{R}(s, \targetpi(s))$ and $\widehat P(s, \targetpi(s), s') = \overline{P}(s, \targetpi(s), s')$. This gives us the following optimization problem: 
\begin{align}
\label{prob.on}
\tag{P2}
&\quad \min_{M, R, P,  \mu^{\targetpi}, \mu^{\neighbor{\targetpi}{s}{a}}} \quad \cost(M, \overline{M}, \costr, \costp, p) \\
\notag
&\quad \mbox{ s.t. } \quad \text{$ \mu^{\targetpi}$ and $P$ satisfy \eqref{eq.stat_dist.from.kernel}}, \\
\notag
&\qquad \quad \quad \forall s, a \ne \targetpi(s): \text{$ \mu^{\neighbor{\targetpi}{s}{a}}$ and $P$ satisfy \eqref{eq.stat_dist.from.kernel}}, \\
\notag
&\qquad \quad \quad \forall s, a \ne \targetpi(s): 
\sum_{s'} \mu^{\targetpi}(s') \cdot R \big (s', \targetpi(s') \big ) \ge
\sum_{s'} \mu^{\neighbor{\targetpi}{s}{a}}(s') \cdot  R \big (s', \neighbor{\targetpi}{s}{a}(s') \big ) + \epsilon,\\
\notag
&\qquad \quad \quad \forall s, a, s': P(s,a,s') \ge \epsP \cdot \overline P(s,a,s'),\\
\notag
&\qquad \quad \quad \forall s, s': P(s, \targetpi(s),s') = \overline P(s, \targetpi(s),s'),\\
\notag
&\qquad \quad \quad \forall s: R(s, \targetpi(s)) = \overline R(s, \targetpi(s)),\\
\notag
&\qquad \quad \quad M = (S, A, R, P, \gamma).
\end{align}

    \item Obtain the sampling MDP $\widehat M$ by solving   \eqref{prob.on}.
    \item \looseness-1 
    Use the sampling MDP $\widehat M$ instead of the environment $\overline M$ during the learning process, i.e., obtain $r_t$ from $\widehat R(s_t, a_t)$ and $s_{t+1} \sim \widehat P(s_t, a_t, .)$ (see Figure~\ref{fig:model.online}). 
\end{itemize}

\subsection{Online Attacks: Theoretical Analysis}
\label{sec.on.analysis}

We now state the formal results that connect the performance of the learning agent---the regret and the number of suboptimal steps---to the average mismatch $\avgmissm(T)$ and the average attack cost $\avgcost(T)$. 
%
Notice that Lemma~\ref{lemma.on.known.nontarget} and Lemma~\ref{lemma.on.subopt.missmatch} directly relate the performance of the learning agent to the average mismatch w.r.t. the policy $\targetpi$. Moreover, we show that the average attack cost can be bounded by the product of two factors: 
one which depends on the learner's performance, and the other that specifies the cost of changing the original MDP $\overline{M}$ to the sampling MDP $\widehat M$, expressed in $\ell_\infty$\emph{-norm}.


More formally, in the average reward criteria with $\gamma=1$, for a learning agent with a bound on the expected regret, we obtain:
\begin{theorem_new}[Average reward criteria, $\gamma=1$]
	\label{theorem.on.regret}
	Let $\widehat M$ be the optimal solution to \eqref{prob.on}. Consider the attack defined by $r_t$ obtained from $\widehat{R}(s_t, a_t)$ and $s_{t + 1} \sim \widehat{P}(s_t, a_t, .)$, and an online learning agent whose expected regret in an MDP $M$ is $\expct{\regret(T, M)}$. The average mismatch of the learner is in expectation upper bounded by
	 $$\expct{\avgmissm(T)} \le  \frac{\widehat{\mu}_{\textnormal{max}}}{\epsilon \cdot T} \cdot \Big(\expct{\regret(T, \widehat M)} + 2\norm{\widehat{V}^{\targetpi}}_\infty\Big).$$
	Furthermore, the average attack cost is in expectation upper bounded by $$\expct{\avgcost(T)} \le 
	 \frac{ \cost(\widehat{M}, \overline{M}, \costr, \costp, \infty)}{T}
	 \cdot 
	 \bigg(
	  \frac{\widehat{\mu}_{\textnormal{max}}}{\epsilon} \cdot \Big(\expct{\regret(T, \widehat{M})} + 2\norm{\widehat{V}^{\targetpi}}_\infty\Big)
	 \bigg)^{1/p}.
	$$
\end{theorem_new}

Similarly, in the discounted reward criteria with $\gamma<1$, for a learning agent with a bound on the number of suboptimal steps, we obtain: 
\begin{theorem_new}[Discounted reward criteria, $\gamma<1$]
	\label{theorem.on.subopt}
	\looseness-1Let $\widehat M$ be the optimal solution to \eqref{prob.on}. Consider the attack defined by $r_t$ obtained from $\widehat{R}(s_t, a_t)$ and $s_{t + 1} \sim \widehat{P}(s_t, a_t, .)$, and an online learning agent whose expected number of suboptimal steps in an MDP $M$ is $\expct{\subopt(T, M, \epsilon')}$. 
	The average mismatch of the learner is in expectation given by
	 $$\expct{\avgmissm(T)} = \frac{1}{T} \cdot \expct{\subopt(T, \widehat M, \epsilon)}.$$
	Furthermore, the average attack cost is in expectation upper bounded by $$\expct{\avgcost(T)} \le 	  \frac{\cost(\widehat{M}, \overline{M}, \costr, \costp, \infty)}{T} \cdot \Big(\expct{\subopt(T, \widehat M, \epsilon)}\Big)^{1/p}.$$
\end{theorem_new}



\noindent~A direct consequence of these theorems is that for a learner with sublinear $\expct{\regret(T, M)}$ (resp. $\expct{\subopt(T, M, \epsilon)}$), both the expected average mismatch and the expected average attack cost will decrease over time, and the rate of decrease depends on the learner's performance.\footnote{For some learning algorithms, guarantees on the learner's performance, i.e., regret or number of suboptimal steps, are true only with high probability: In that case, one would modify the results in Theorem~\ref{theorem.on.subopt} and Theorem~\ref{theorem.on.regret} accordingly.}
Note that, while we considered $\ell_p$ norms with $p \ge 1$ to define the attack cost, the above results can be generalized to include the case of $p=0$. For $p=0$, the expected average cost is equivalent to the expected number of average mismatches, and the same upper bound applies.
\subsection{Online Attacks: Efficiency of Solving the Problem}\label{sec.on.problem.refromulation}


In Section \ref{sec.on.problem}, we outlined a template for attacking an online learner that uses optimization problem \eqref{prob.on} as its subroutine. In this subsection, we show that \eqref{prob.on} can be reformulated as tractable convex program with linear constraints, which increases the computational efficiency of the proposed attack, and makes it more scalable. 

Observe that the first three constraints in optimization problem \ref{prob.on} are quadratic constraints. Since the attack does not change the transitions associated to $\targetpi$, i.e., $P(s, \targetpi(s), s') = \overline{P}(s, \targetpi(s), s')$, we have $\mu^{\targetpi} = \overline{\mu}^{\targetpi}$ and is no longer a variable in the optimization problem (i.e., it is precomputed).

To tackle the 2nd and 3rd quadratic constraints, we use two key ideas which will enable us to express these constraints with a linear constraint. The first key idea is to relate the scores of the target policy and its neighbor policies, i.e., $\rho^{\targetpi}$ and $\rho^{\neighbor{\targetpi}{s}{a}}$, to their the target policy's $Q$-values. By Corollary \ref{gain_diff_neighbor} in Appendix \ref{appendix.off.general}, we know that
\begin{gather}
\label{equation_target_gap_in_q}
	\rho^{\targetpi} - \rho^{\neighbor{\targetpi}{s}{a}} = 
	\mu^{\neighbor{\targetpi}{s}{a}}(s) \cdot \big(V^{\targetpi}(s) - Q^{\targetpi}(s,a)\big).
\end{gather}
%
%
This identity enables us to rewrite the third constraint in optimization problem \eqref{prob.on} so that there are no quadratic terms of the form $\mu^{\neighbor{\targetpi}{s}{a}}(s') \cdot R(s, \neighbor{\targetpi}{s}{a}(s'))$. 
In particular, using Bellman equations \eqref{equation.bellman}, equation \eqref{equation_target_gap_in_q}, and the fact that $V^{\targetpi} = \overline{V}^{\targetpi}$ and $\rho^{\targetpi} = \overline{\rho}^{\targetpi}$, we can rewrite the 3rd  constraint in \eqref{prob.on} as
\begin{align}
\label{eq.v.constraint}
\forall s, a \ne \targetpi(s): \quad \overline{V}^{\targetpi}(s) - R(s, a) + \overline{\rho}^{\targetpi} - \gamma \sum_{s'}P(s, a, s') \cdot  \overline{V}^{\targetpi}(s') \ge  \frac{\epsilon}{\mu^{\neighbor{\targetpi}{s}{a}}(s)}.
\end{align}
%
%
The only remaining nonlinear part in the modified constraint, i.e., equation \eqref{eq.v.constraint}, is $\frac{1}{\mu^{\neighbor{\targetpi}{s}{a}}(s)}$. Now, we use the second key idea, which we also enable us to remove the 2nd constraint. We rewrite $\frac{1}{\mu^{\neighbor{\targetpi}{s}{a}}(s)}$ in terms of reach times $\overline{T}^{\targetpi}$, which can be precomputed. 
More precisely, in Lemma \ref{lemma.mu.equality.bound} from Appendix~\ref{appendix.off.joint}, we show that
	\begin{align*}
	\frac{1}{\mu^{\neighbor{\pi}{s}{a}}(s)} =  \frac
	{1 + \gamma \cdot  \sum_{s'} P(s, a, s')\cdot \overline{T}^{\pi}(s', s)}
	{1 - (1 - \gamma) \cdot  \sum_{s'}d_0(s') \cdot  \overline{T}^{\pi}(s', s)}.
	\end{align*}{}
%
Using this result, we can rewrite the equation \eqref{eq.v.constraint} as 
\begin{align*}
\forall s, a \ne \targetpi(s): 
\overline{V}^{\targetpi}(s) - R(s, a) + \overline{\rho}^{\targetpi} \nonumber
- \gamma \cdot \sum_{s'}P(s, a, s')\cdot \bigg(\overline{V}^{\targetpi}(s') + \frac{ \epsilon}{\overline{\eta}(s)}\cdot \overline{T}^{\targetpi}(s', s)\bigg) \ge  \frac{\epsilon}{\overline \eta (s)},
\end{align*}
where $\overline \eta(s) = 1 - (1 - \gamma)\sum_{s'}d_0(s')\overline{T}^{\targetpi}(s', s)$.
Note that variable $\mu^{\neighbor{\targetpi}{s}{a}}$ is no longer needed in the optimization problem, which makes the 2nd constraint redundant. 
The new formulation of optimization problem \eqref{prob.on} is therefore given by the following:
\begin{align}
\label{prob.on.reformulated}
\tag{P2'}
&\quad \min_{M, R, P} \quad \cost(M, \overline{M}, \costr, \costp, p) \\
\notag
&\qquad \quad \quad \forall s, a \ne \targetpi(s): \\
\notag
&\qquad \quad \quad \quad \quad \overline{V}^{\targetpi}(s) - R(s, a) + \overline{\rho}^{\targetpi}  - \gamma\cdot\sum_{s'}P(s, a, s')\cdot\bigg(\overline{V}^{\targetpi}(s') + \frac{ \epsilon}{\overline{\eta}(s)}\cdot\overline{T}^{\targetpi}(s', s)\bigg) \ge  \frac{\epsilon}{\overline \eta (s)},\\
\notag
&\qquad \quad \quad \forall s, a, s': P(s,a,s') \ge \epsP \cdot \overline P(s,a,s'),\\
\notag
&\qquad \quad \quad \forall s, s': P(s, \targetpi(s),s') = \overline P(s, \targetpi(s),s'),\\
\notag
&\qquad \quad \quad \forall s: R(s, \targetpi(s)) = \overline R(s, \targetpi(s)),\\
\notag
&\qquad \quad \quad M = (S, A, R, P, \gamma).
\end{align}
%
%
Since $\overline{\rho}^{\targetpi}$, $\overline{V}^{\targetpi}$, and $\overline T^{\targetpi}$ can be efficiently precomputed based on MDP $\overline{M}$, and new constraints are linear in the optimization variables, optimization problem \eqref{prob.on.reformulated} is convex and can be efficiently solved. 
\begin{proposition_new}
	\label{theorem.on.reformulate}
	Problem \eqref{prob.on.reformulated} is a reformulation of problem \eqref{prob.on} and is a convex optimization problem with linear constraints.
\end{proposition_new}

We conclude this section by noting that optimization problem \eqref{prob.on.reformulated} can be solved separately for each state-action pair $(s, a \ne \targetpi(s))$. Namely, the first constraint in \eqref{prob.on.reformulated} only involves parameters of state-action pair $(s, a)$, while the cost function is the $\ell_p$-norm of a vector whose each component only depends on the rewards and transitions of one state-action pair $(s, a)$. Hence, \eqref{prob.on.reformulated} can be broken into $|S| \cdot (|A|-1)$ independent problems.

\section{Numerical Simulations}\label{sec.experiments}
In this section, we perform numerical simulations and empirically investigate the effectiveness of the proposed attacks on two different environments.  For the reproducibility of experimental
results and facilitating research in this area, the source code of our implementation is publicly available.\footnote{Code: \url{https://machineteaching.mpi-sws.org/files/jmlr2020_rl-policy-teaching_code.zip}.}

\subsection{Environments}\label{sec.experiments.environments}

\begin{figure*}[t!]
\centering
\begin{minipage}{0.38\textwidth}
\centering
	\includegraphics[width=0.92\linewidth]{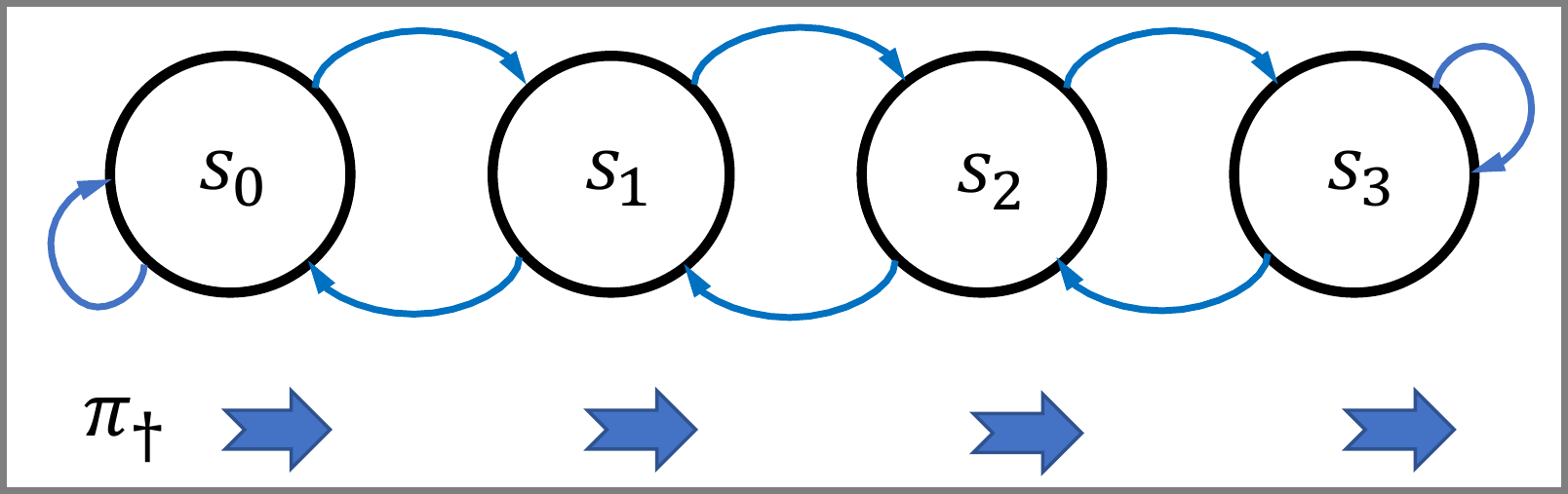}
    \vspace{12mm}	
	\caption{Chain environment with $|S| = 4$ states and $|A|=2$ actions.}
	\label{fig:environment.chain}
\end{minipage}
\quad
\begin{minipage}{0.58\textwidth}
\centering
\includegraphics[width=1\linewidth]{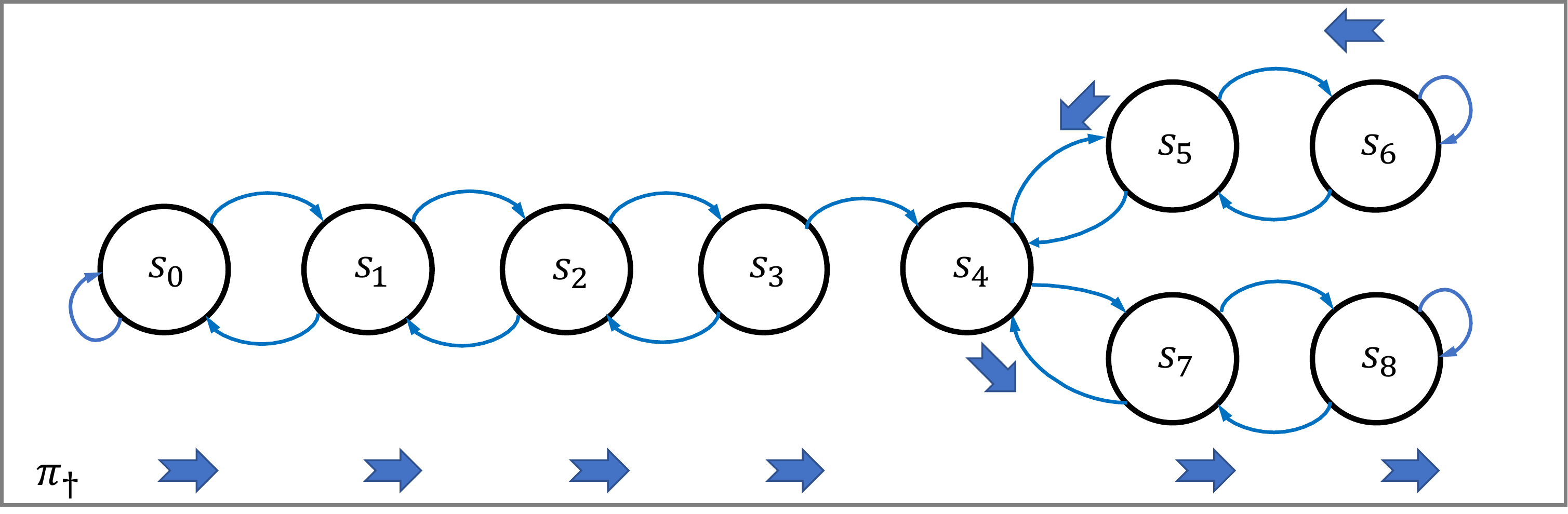}
	\caption{Navigation environment with $|S| = 9$ states and $|A|=2$ actions.}
	\label{fig:environment.grid}
\end{minipage}
\vspace{4mm}
\end{figure*}

The first environment we consider is a \emph{chain} environment represented as an MDP with four states and two actions, see Figure~\ref{fig:environment.chain}. Even though simple, this environment provides a very rich and an intuitive problem setting to validate the theoretical statements and understand the effectiveness of the attacks by varying different parameters. We will also vary the number of states in the MDP to check the efficiency of solving different optimization problems, and report run times.  The second environment we consider is a \emph{navigation} environment represented as an MDP with nine states and two actions per state, see Figure~\ref{fig:environment.grid}. This environment is inspired by a navigation task and is slightly more complex than the environment in Figure~\ref{fig:environment.chain}.  Below, we provide specific details of these two environments. 

\paragraph{Details of the chain environment.} 
The environment has $|S| = 4$ states and $|A|=2$ actions given by $\{\texttt{left}, \texttt{right}\}$. The original reward function $\overline{R}$ is action independent and has the following values: $s_1$ and $s_2$ are rewarding states with $\overline{R}(s_1,.) =  \overline{R}(s_2,.) = 0.5$, state $s_3$ has negative reward of $\overline{R}(s_3,.) = -0.5$, and the reward of the state $s_0$ given by $\overline{R}(s_0,.)$  will be varied in experiments. With probability $0.9$, the actions succeed in navigating the agent to left or right as shown on arrows; with probability $0.1$ the agent's next state is sampled randomly from the set $S$. The target policy $\targetpi$ is to take \texttt{right} action in all states as shown in Figure \ref{fig:environment.chain}. 

\paragraph{Details of the navigation environment.}
The environment has $|S| = 9$ states and $|A|=2$ actions per state. The original reward function $\overline{R}$ is action independent and has the following values: $\overline{R}(s_1,.) = \overline{R}(s_2,.) = \overline{R}(s_3,.) = -2.5$,  $\overline{R}(s_4,.) = \overline{R}(s_5,.) = 1.0$, $\overline{R}(s_6,.) = \overline{R}(s_7,.) = \overline{R}(s_8,.) = 0$, and the reward of the state $s_0$ given by $\overline{R}(s_0,.)$ will be varied in experiments. With probability $0.9$, the actions succeed in navigating the agent as shown on arrows; with probability $0.1$ the agent's next state is sampled randomly from the set $S$. The target policy $\targetpi$ is to take actions as shown with bold arrows in Figure \ref{fig:environment.grid}.


\vspace{2mm}
\subsection{Attacks in the Offline Setting: Setup and Results}\label{sec.experiments.offline}

\paragraph{Attack strategies.} 
%
For the offline setting, we compare the performance of our attack strategy (\JAttack) with three different baseline strategies (\NTJAttack, \RAttack, \DAttack) as discussed below:
\begin{enumerate}
	\item \JAttack: joint rewards and transitions attack using ($\widehat{R}, \widehat{P}$) obtained as a solution to the problem \eqref{prob.off}.
    \item \NTJAttack: joint rewards and transitions attack using ($\widehat{R}, \widehat{P}$) obtained as a solution to the problem  \eqref{prob.on}; here, \textsc{NT-} prefix is used to highlight that \emph{non-target only} manipulations are allowed.
    \item \RAttack: rewards only attack obtained as a solution to the problem \eqref{prob.off} when $\widehat{P} := \overline{P}$ (alternatively, by taking the limit of $\costp$  to infinity in the problem).
    \item \DAttack: transitions only attack obtained as a solution to the problem \eqref{prob.off} when $\widehat{R} := \overline{R}$ (alternatively, by taking the limit of $\costr$  to infinity in the problem).
\end{enumerate}



 
We set $p = \infty$ (i.e., $\ell_\infty$\emph{-norm}) in the objective when solving different attack problems.
%
We note that optimal solutions for the problems corresponding to \RAttack~and \NTJAttack~can be computed efficiently using standard optimization techniques (also, refer to discussions in Section~\ref{sec.off.efficiency} and Section~\ref{sec.on.problem.refromulation}). Problems corresponding to \JAttack~and \DAttack~are computationally more challenging, and we provide a simple yet effective approach towards finding an approximate solution---specific implementation details are provided in Appendix~\ref{appendix.sec.experiments}.
For more detailed results and analysis of the rewards only and transitions only attacks (\RAttack~and \DAttack), we refer the reader to the earlier version of the paper~\citep{rakhsha2020policy}.

\paragraph{Experimental setup and parameter choices.} For all the experiments, we set $\costr=3$, $\costp=1$, and use $\ell_\infty$-norm in the measure of the attack cost (see Section~\ref{sec.formulation.offline}). The regularity parameter $\epsP$ in the problems \eqref{prob.off}~and~\eqref{prob.on} is set to be $0.0001$. In the experiments, we vary $\overline{R}(s_0, .) \in [-5, 5]$ and vary $\epsilon$ margin $\in [0, 1]$ for the $\targetpi$ policy. The results are reported as an average of $10$ runs.  For the offline setting, we only report results for the discounted reward criteria with $\gamma=0.99$; the results for the average reward criteria are very similar to the ones reported here. For the chain environment, we also vary the number of states $|S|$ and report run times for solving different attack problems.

\begin{figure*}[t!]
\centering
	\begin{subfigure}[b]{0.35\textwidth}
	   \centering
		\includegraphics[width=1\linewidth]{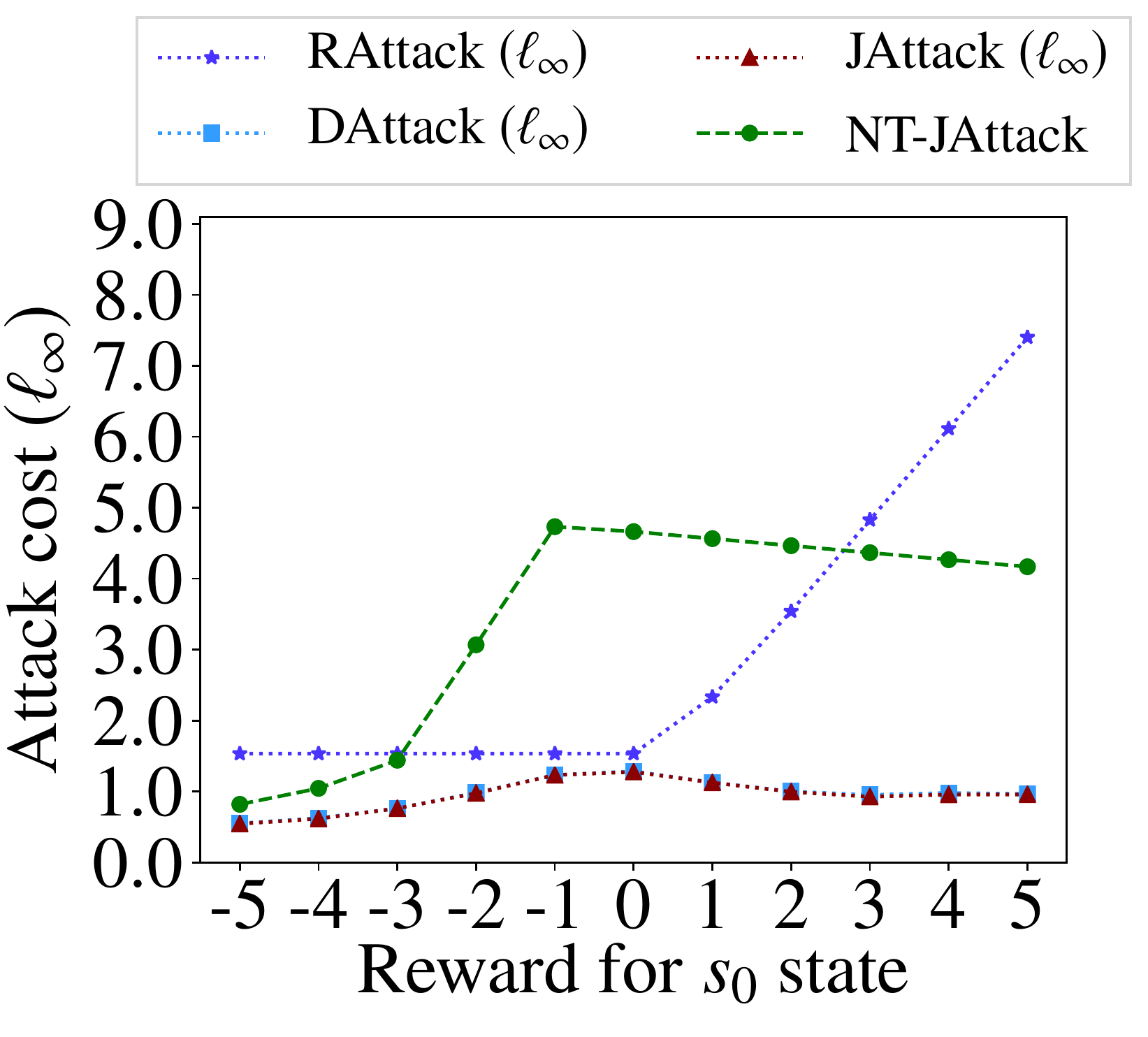}
		\caption{Vary $\overline{R}(s_0, .)$}
		\label{fig:results.offline.chain.1}
	\end{subfigure}
	\qquad \qquad	
	\begin{subfigure}[b]{0.35\textwidth}
	    \centering
		\includegraphics[width=1\linewidth]{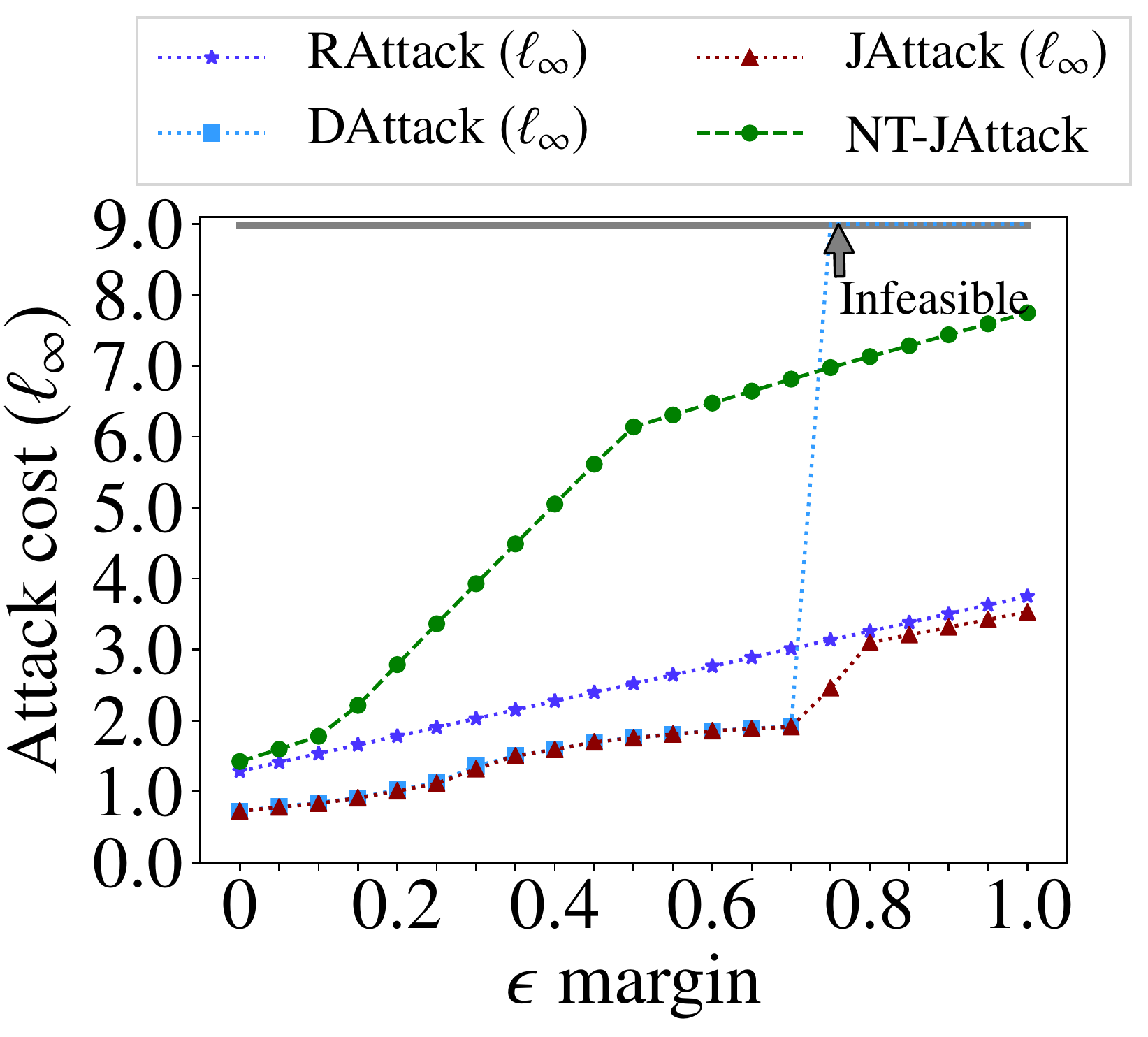}
		\caption{Vary $\epsilon$ margin}
		\label{fig:results.offline.chain.2}
	\end{subfigure}		
	\vspace{-2mm}	
   \caption{(\textbf{Chain environment}) Results for poisoning attacks in the offline setting from Section~\ref{sec.offlineattacks}. (\textbf{a}) shows results when we vary reward $\overline{R}(s_0, .)$ and (\textbf{b}) shows results when we vary $\epsilon$ margin.}
	\label{fig:results.offline.chain}
	\vspace{2mm}
\end{figure*}

\begin{figure*}[t!]
\centering	
	\begin{subfigure}[b]{0.35\textwidth}
	    \centering
		\includegraphics[width=1\linewidth]{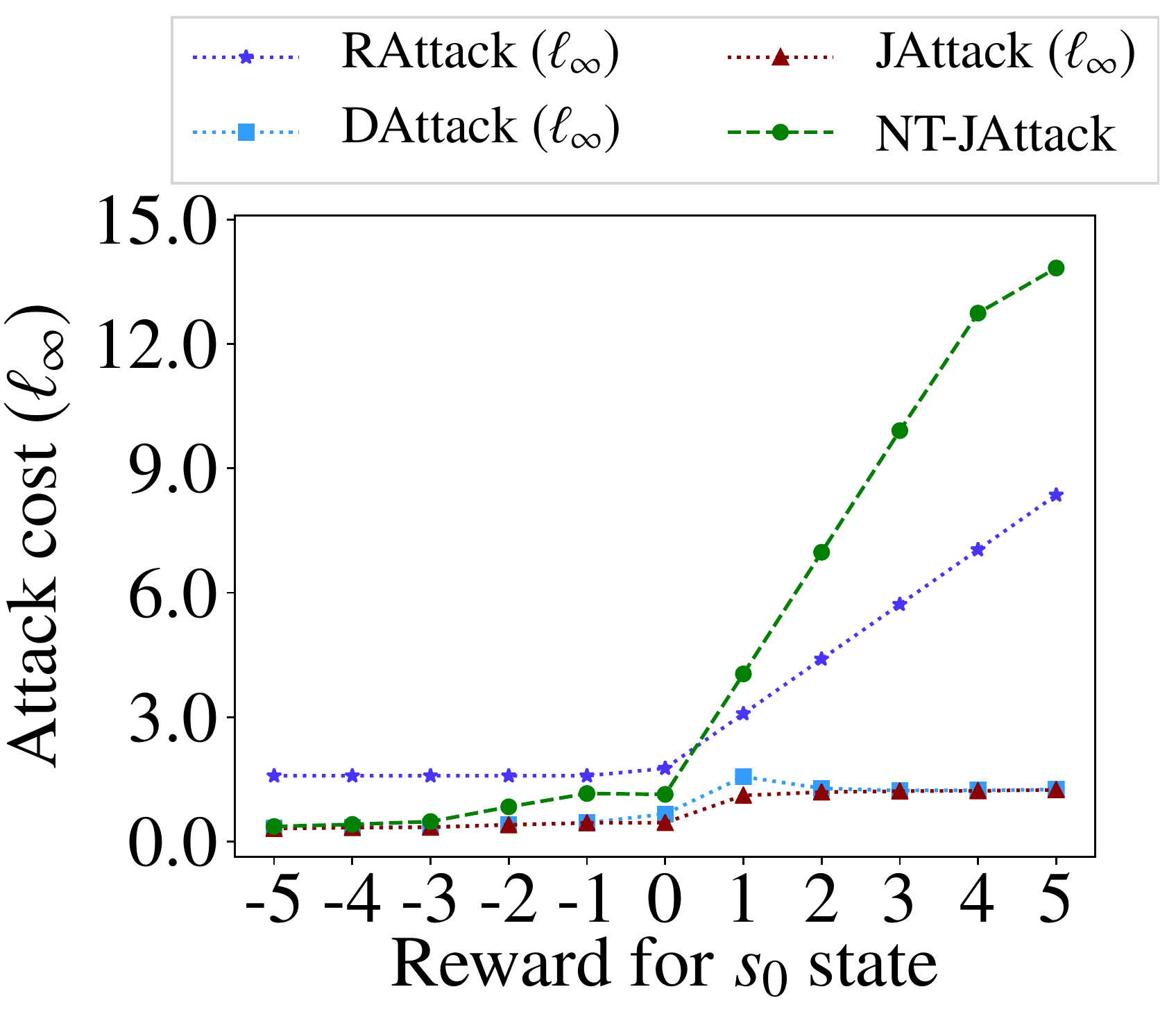}
		\caption{Vary $\overline{R}(s_0, .)$}
		\label{fig:results.offline.grid.1}
	\end{subfigure}
	\qquad \qquad
	\begin{subfigure}[b]{0.35\textwidth}
	    \centering
		\includegraphics[width=1\linewidth]{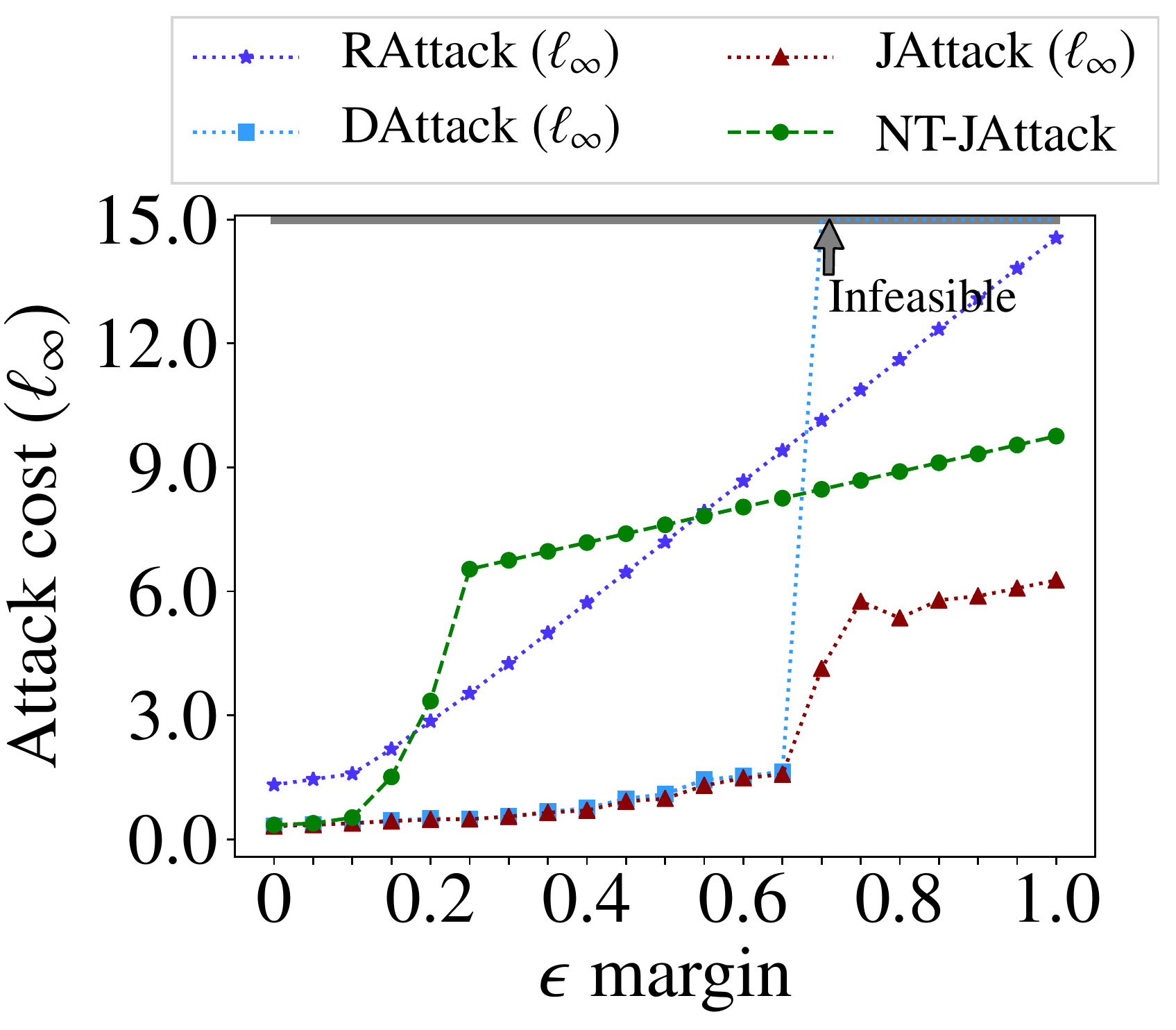}
		\caption{Vary $\epsilon$ margin}
		\label{fig:results.offline.grid.2}
	\end{subfigure}
	\vspace{-2mm}	
	   \caption{(\textbf{Navigation environment}) Results for poisoning attacks in the offline setting from Section~\ref{sec.offlineattacks}. (\textbf{a}) shows results when we vary reward $\overline{R}(s_0, .)$ and (\textbf{b}) shows results when we vary $\epsilon$ margin.}
	\label{fig:results.offline.grid}
	\vspace{2mm}
\end{figure*}

\paragraph{Results.}  Figure~\ref{fig:results.offline.chain} reports results for the chain environment (with $|S|=4$) and  Figure~\ref{fig:results.offline.grid} reports results for the navigation environment. The key takeaways are same for both the environments, and we want to highlight three points here. First, as we increase the desired $\epsilon$ margin, the attack problem becomes more difficult. While the attacks that allow reward poisoning  (\JAttack, \NTJAttack, \RAttack) are always feasible though with increasing attack cost,  it becomes infeasible to do transitions only poisoning attack (\DAttack) (e.g., in Figure~\ref{fig:results.offline.chain}, \DAttack~attack is not possible for $\epsilon > 0.75$). Second, the plots show that joint attack (\JAttack) can have much lower cost compared to reward only attack (\RAttack) or transitions only attack (\DAttack). Third, the plots also show that our joint attack strategy (\JAttack) has much lower cost compared to the non-target only attack strategy (\NTJAttack). Finally, to check the efficiency of solving above mentioned attack problems, we vary the number of states in the chain environment and report the run times in Table \ref{table:runtimes.chain}.

\begin{table}[t!]
  \centering
    \begin{tabular}{|r|r|r|r|r|r|r|r|}
		\hline
	    \backslashbox{Attacks}{$|S|$} & $4$ & $10$ & $20$ & $30$ & $50$ & $70$ &$100$  \\
		\hline
		\RAttack & 0.01s & 0.02s & 0.04s & 0.06s & 0.14s & 0.29s & 0.61s \\
		\DAttack & 3.09s & 7.46s & 14.98s & 24.73s & 46.02s & 77.57s & 126.97s\\
		\NTJAttack & 0.06s & 0.11s & 0.22s & 0.34s & 0.60s & 0.83s & 1.27s\\
		\JAttack & 8.35s & 20.52s & 42.45s & 64.98s & 116.01s & 180.36s & 273.20s\\
		\hline
    \end{tabular}
    \vspace{-2mm}
    \caption{(\textbf{Chain environment}) Run times for solving different attack problems as we vary the number of states $|S|$. Numbers are reported in seconds and are based on an average of $5$ runs for each setting.}
    \label{table:runtimes.chain}
	\vspace{2mm}    
\end{table}
\vspace{8mm}
\subsection{Attacks in the Online Setting: Setup and Results}\label{sec.experiments.online}

\paragraph{Attack strategies.} 
%
For the online setting, we compare the performance of our attack strategy (\NTJAttack) with two different baseline strategies (\JAttack, \NoAttack) as discussed below:
\begin{enumerate}
    \item \NTJAttack: joint rewards and transitions attack using ($\widehat{R}, \widehat{P}$) obtained as a solution to the problem  \eqref{prob.on}; here, \textsc{NT-} prefix is used to highlight that \emph{non-target only} manipulations are allowed.
	\item \JAttack: joint rewards and transitions attack using ($\widehat{R}, \widehat{P}$) obtained as a solution to the problem \eqref{prob.off}.
    \item \NoAttack: a default setting without adversary denoted as \NoAttack~where environment feedback is sampled from the original MDP $\overline{M}$.
\end{enumerate}

The implementation details for \NTJAttack~and \JAttack~are discussed above in Section~\ref{sec.experiments.offline} and Appendix~\ref{appendix.sec.experiments}.

\paragraph{Experimental setup and parameter choices.} For all the experiments, we set $\costr=3$, $\costp=1$. The regularity parameter $\epsP$ in the problems \eqref{prob.off}~and~\eqref{prob.on} is set to be $0.0001$. In the experiments, we fix $\overline{R}(s_0, .)=-2.5$ and $\epsilon=0.1$ margin for the $\targetpi$ policy.  We plot the measure of the attacker's achieved goal in terms of \avgmissm~and attacker's cost in terms of \avgcost~for $\ell_1$\emph{-norm} measured over time $t$ (see Section~\ref{sec.formulation.online}).  The results are reported as an average of $20$ runs. We separately run experiments for the average reward optimality criteria (with $\gamma = 1$) and the discounted reward optimality criteria (with $\gamma = 0.99$). For the  average reward  criteria, we consider an RL agent implementing the UCRL learning algorithm~\cite{auer2007logarithmic}. For the discounted reward criteria, we consider an RL agent implementing Q-learning with an exploration parameter set to $0.001$ \cite{even2003learning}. For further details about the RL agents, we refer the reader to Section~\ref{subsec.online_agent} and Appendix~\ref{appendix_background}. For both the settings, the attacker does not use any knowledge of the agent's learning algorithm.

\begin{figure*}[t!]
\centering
	\begin{subfigure}[b]{0.35\textwidth}
	   \centering
		\includegraphics[width=1\linewidth]{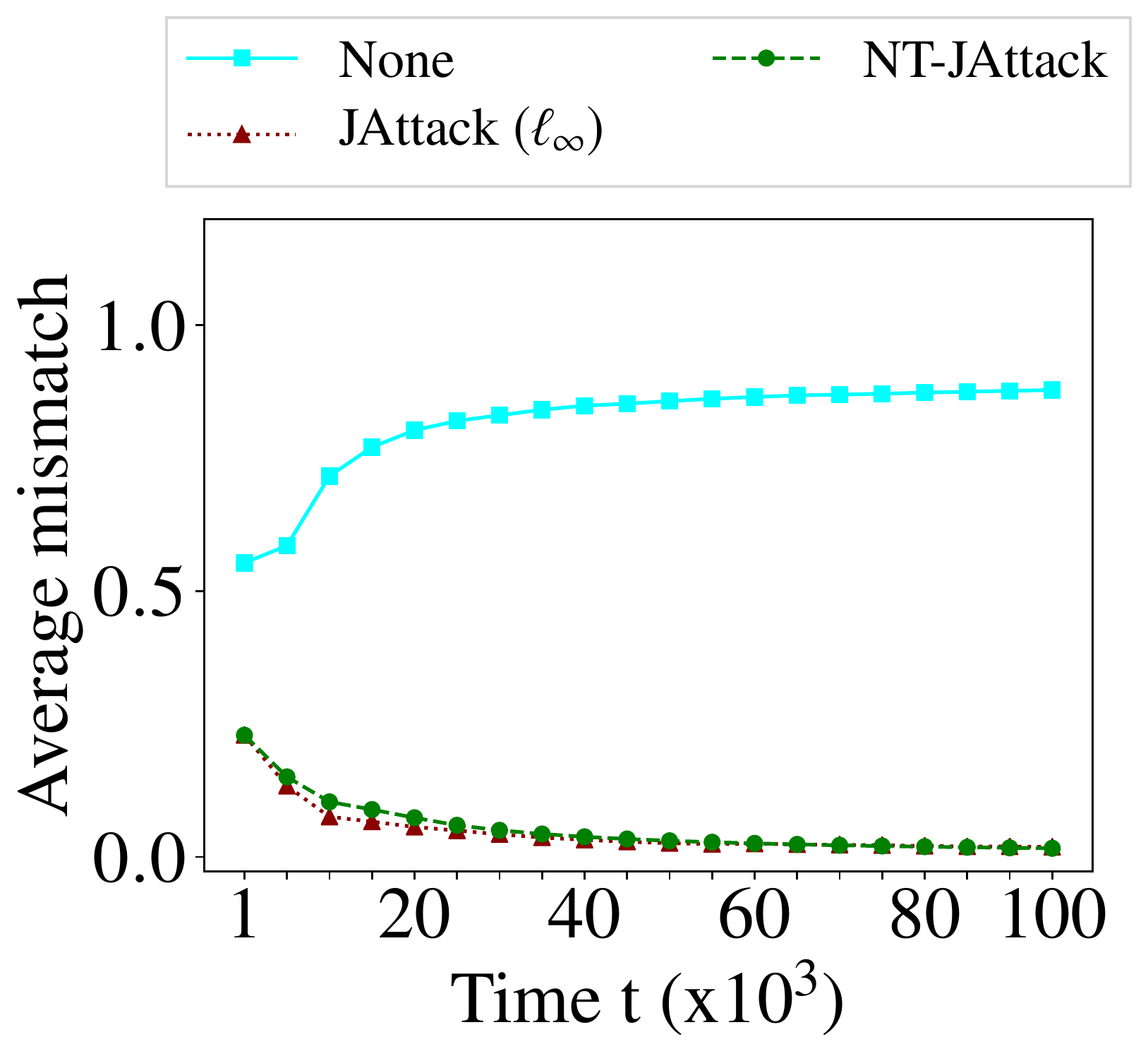}
		\caption{\avgmissm: $\gamma=1$}
		\label{fig:results.online.chain.1}
	\end{subfigure}
	\qquad \qquad
	\begin{subfigure}[b]{0.35\textwidth}
	    \centering
		\includegraphics[width=1\linewidth]{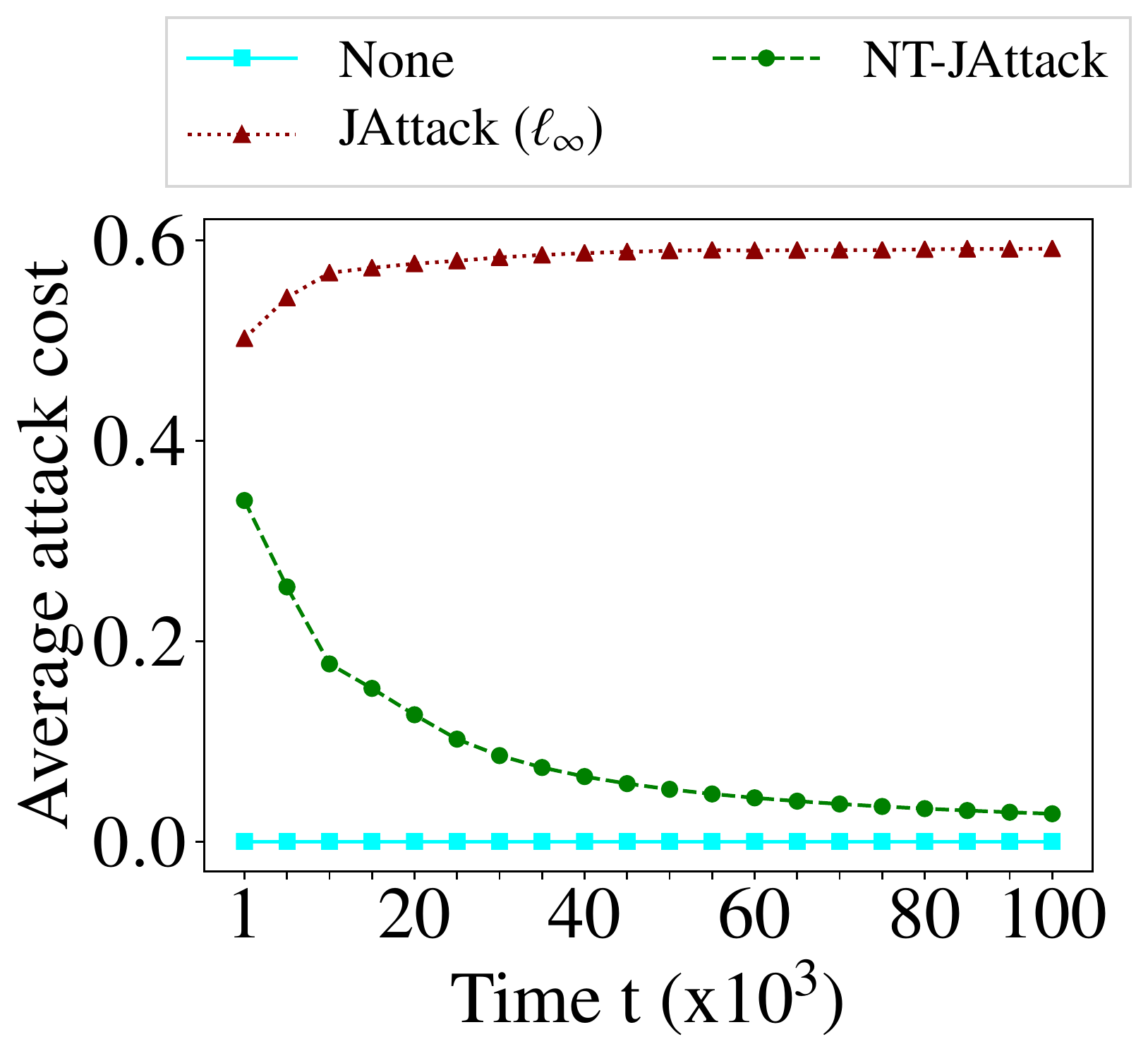}
		\caption{\avgcost: $\gamma=1$}
		\label{fig:results.online.chain.2}
	\end{subfigure}
	\\
	\vspace{2mm}
	\begin{subfigure}[b]{0.35\textwidth}
	    \centering
		\includegraphics[width=1\linewidth]{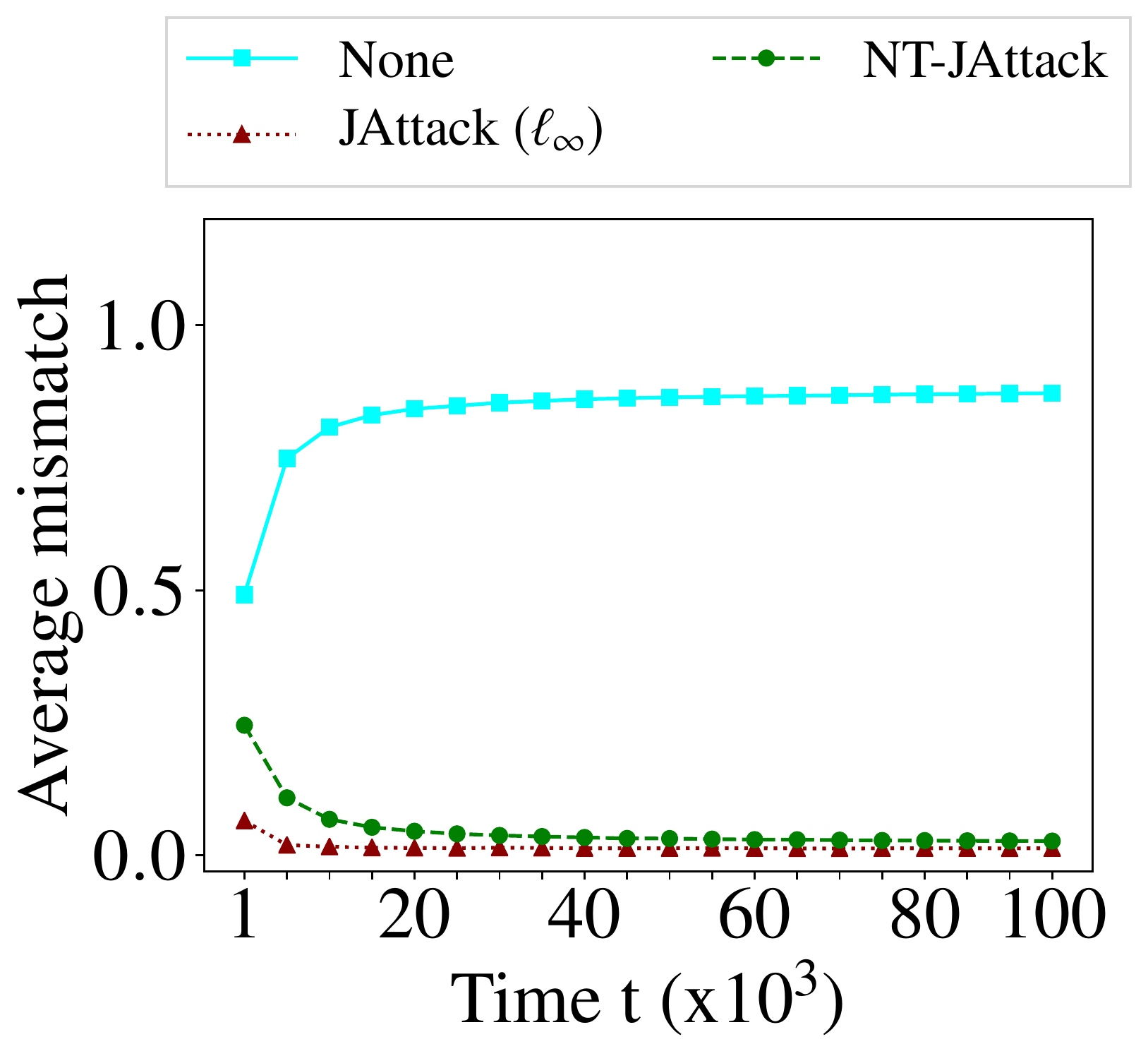}
		\caption{\avgmissm: $\gamma=0.99$}
		\label{fig:results.online.chain.3}
	\end{subfigure}
	\qquad \qquad		
	\begin{subfigure}[b]{0.35\textwidth}
	    \centering
		\includegraphics[width=1\linewidth]{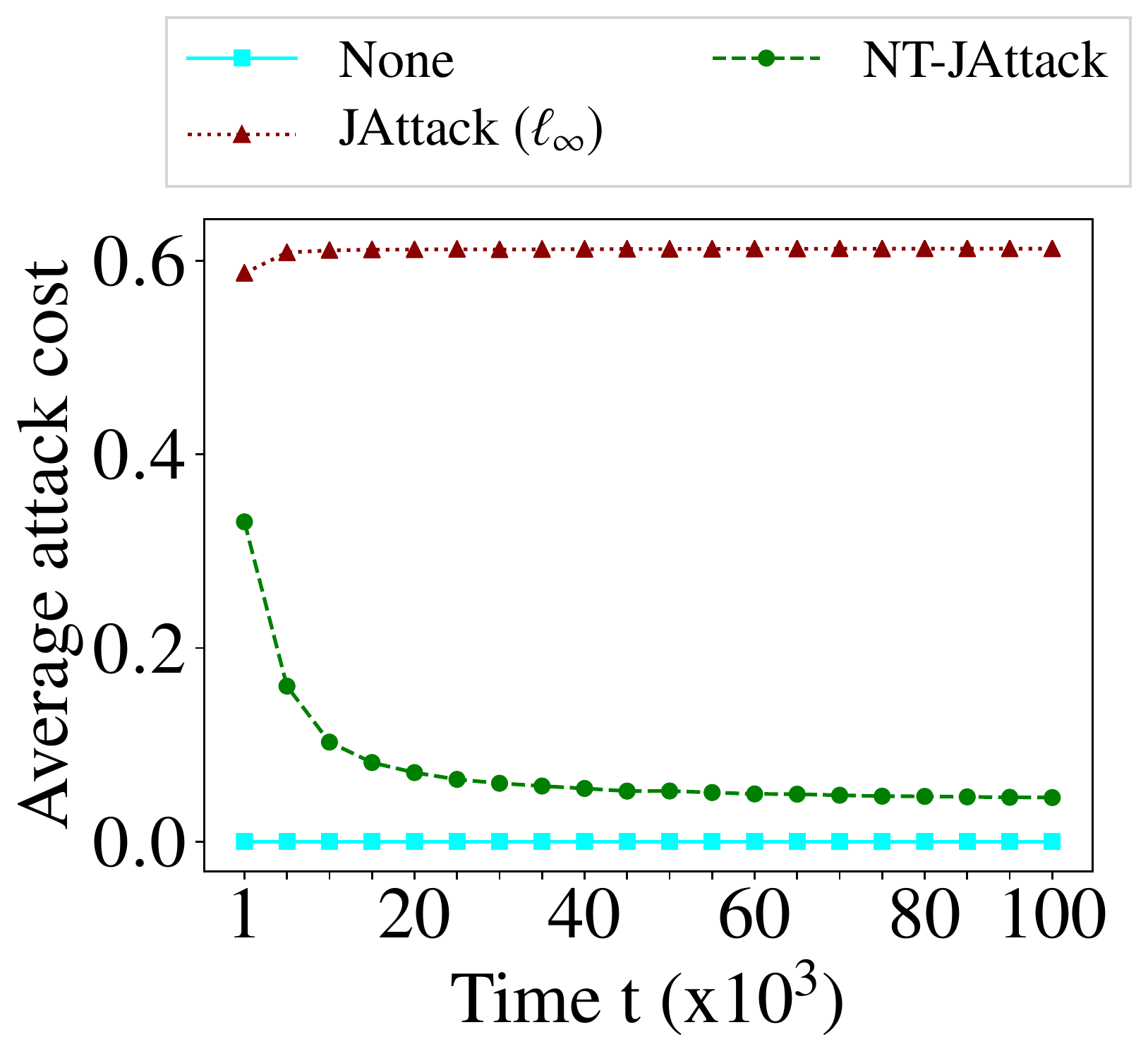}
		\caption{\avgcost: $\gamma=0.99$}
		\label{fig:results.online.chain.4}
	\end{subfigure}			
	\vspace{-2mm}	
   \caption{(\textbf{Chain environment}) Results for poisoning attacks in the online setting from Section~\ref{sec.onlineattacks}. (\textbf{a}, \textbf{b}) plots show results for the average reward criteria ($\gamma=1$) with UCRL as the agent's learning algorithm. (\textbf{c}, \textbf{d}) plots show results for the discounted reward criteria ($\gamma=0.99$) with Q-learning as the agent's learning algorithm.
   }
	\label{fig:results.online.chain}
\end{figure*}
   

\begin{figure*}[t!]
\centering
	\begin{subfigure}[b]{0.35\textwidth}
	   \centering
		\includegraphics[width=1\linewidth]{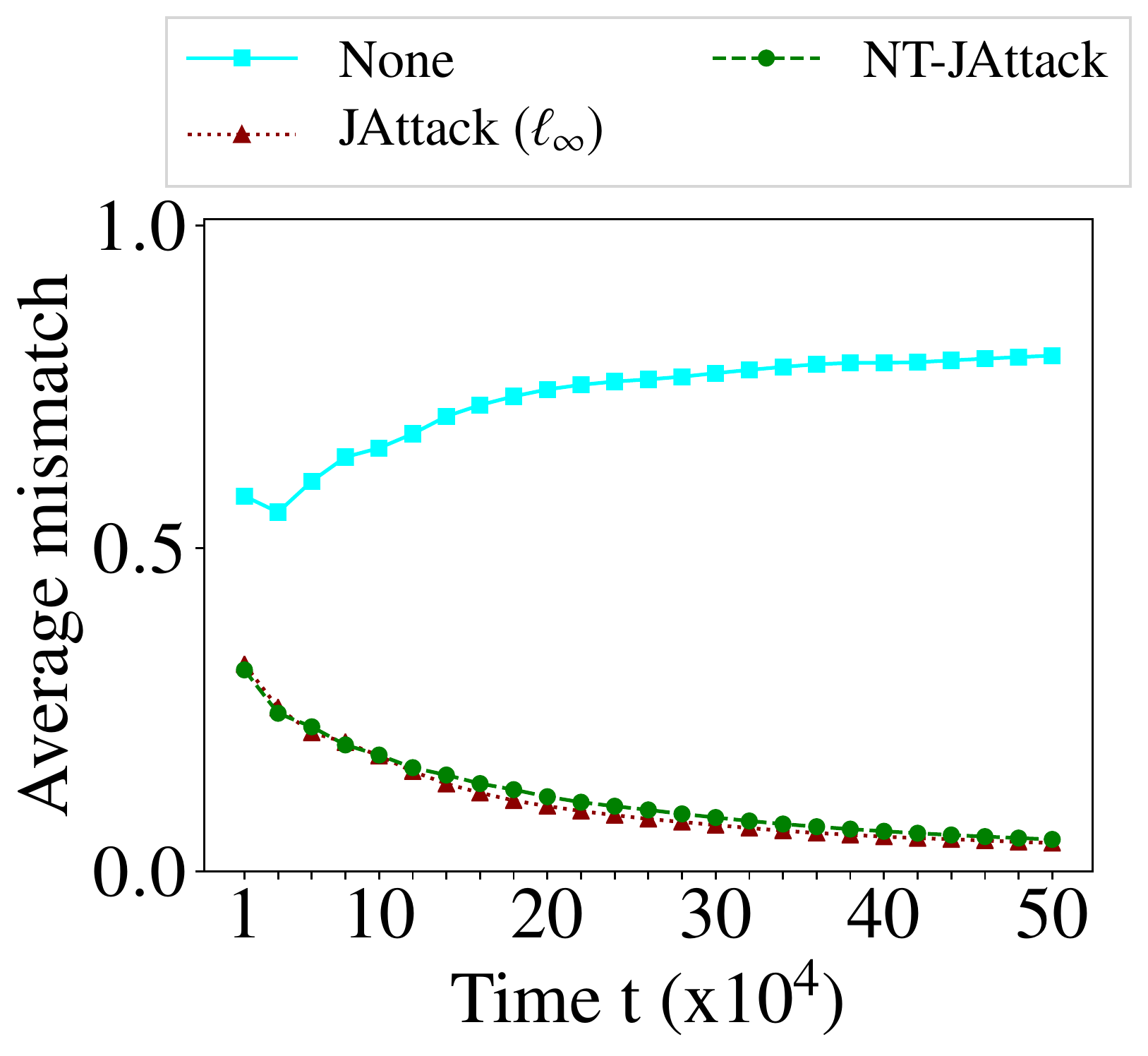}
		\caption{\avgmissm: $\gamma=1$}
		\label{fig:results.online.grid.1}
	\end{subfigure}
	\qquad \qquad
	\begin{subfigure}[b]{0.35\textwidth}
	    \centering
		\includegraphics[width=1\linewidth]{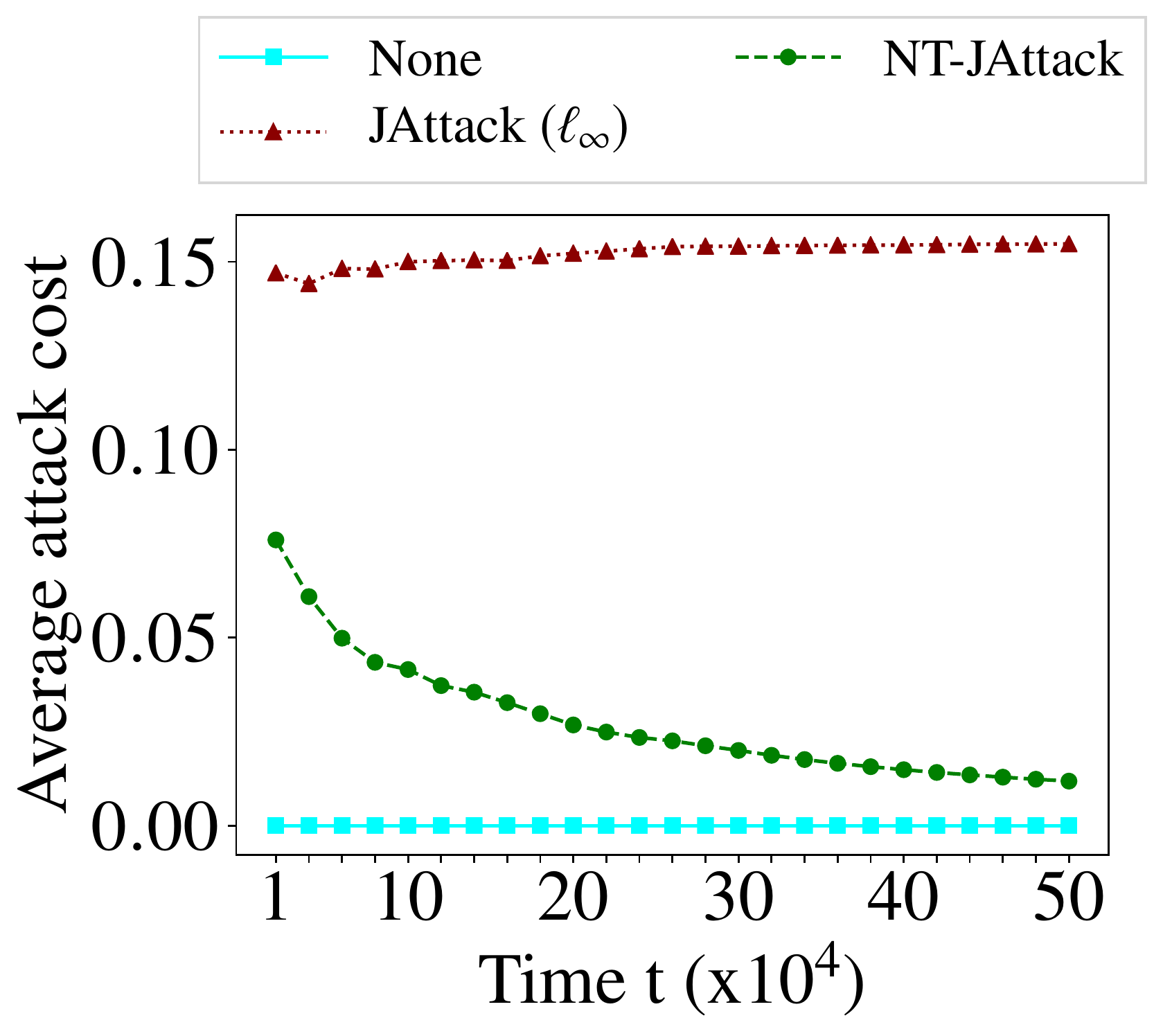}
		\caption{\avgcost: $\gamma=1$}
		\label{fig:results.online.grid.2}
	\end{subfigure}
	\\
	\vspace{2mm}
	\begin{subfigure}[b]{0.35\textwidth}
	    \centering
		\includegraphics[width=1\linewidth]{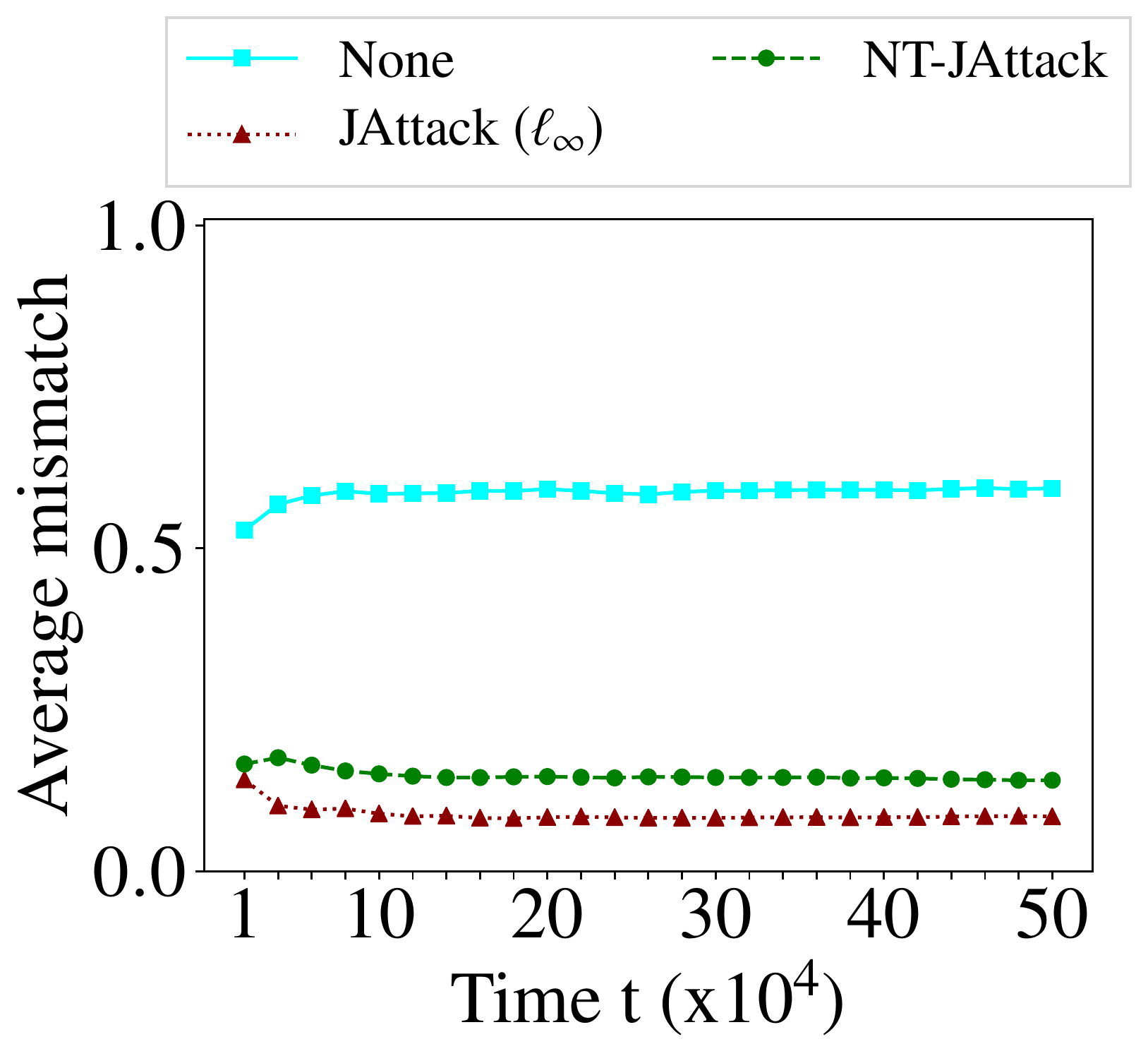}		
		\caption{\avgmissm: $\gamma=0.99$}
		\label{fig:results.online.grid.3}
	\end{subfigure}		
	\qquad \qquad	
	\begin{subfigure}[b]{0.35\textwidth}
	    \centering
		\includegraphics[width=1\linewidth]{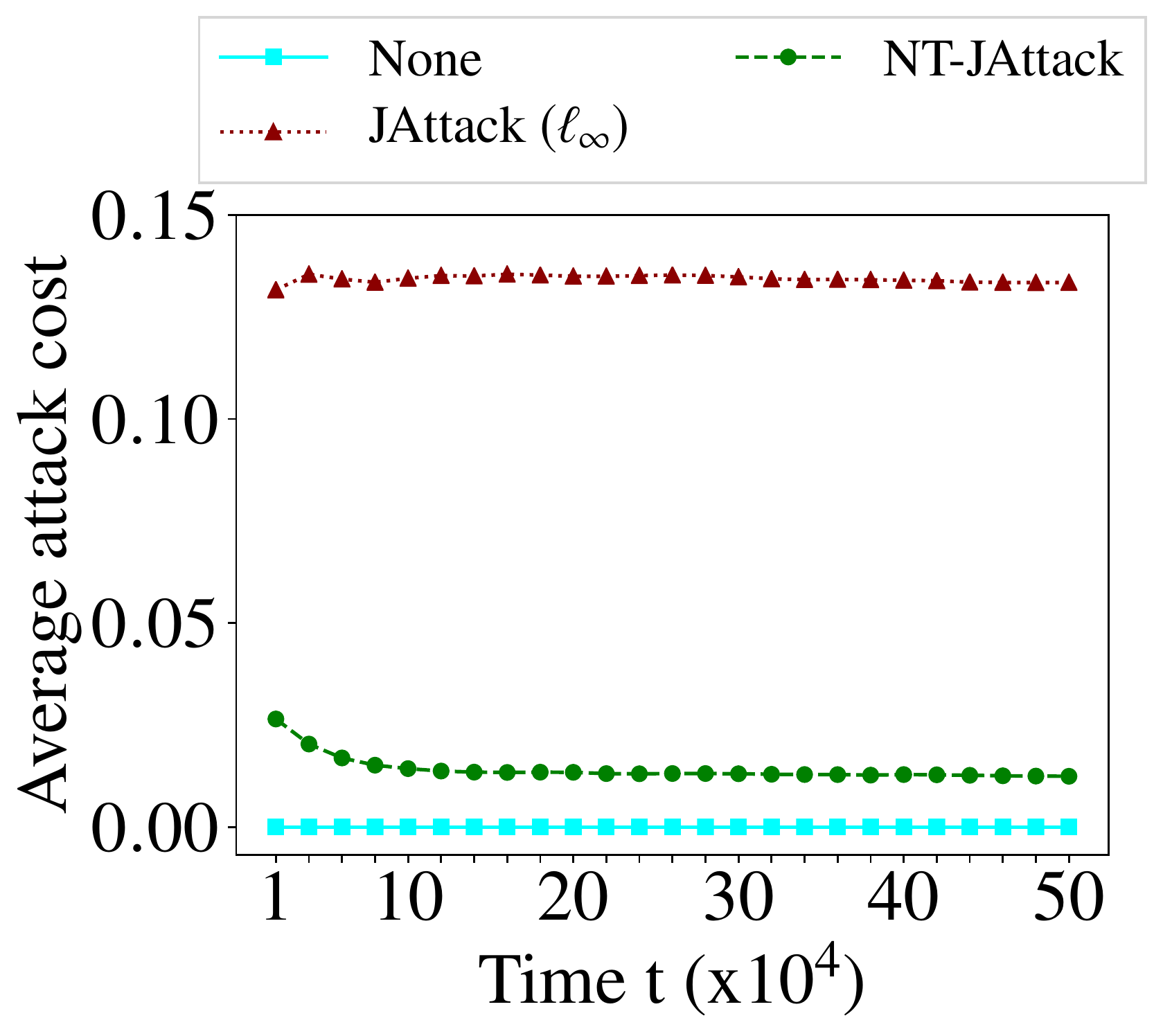}		
		\caption{\avgcost: $\gamma=0.99$}
		\label{fig:results.online.grid.4}
	\end{subfigure}			
	\vspace{-2mm}	
   \caption{(\textbf{Navigation environment}) Results for poisoning attacks in the online setting from Section~\ref{sec.onlineattacks}. (\textbf{a}, \textbf{b}) plots show results for the average reward criteria ($\gamma=1$) with UCRL as the agent's learning algorithm. (\textbf{c}, \textbf{d}) plots show results for the discounted reward criteria ($\gamma=0.99$) with Q-learning as the agent's learning algorithm.
   }	
	\label{fig:results.online.grid}
\end{figure*}

\paragraph{Results.}  Figure~\ref{fig:results.online.chain} reports results for the chain environment (with $|S|=4$) and  Figure~\ref{fig:results.online.grid} reports results for the navigation environment. The key takeaways are same for both the environments, and we want to highlight two points here.  First, the results in Figures~\ref{fig:results.online.chain}~and~\ref{fig:results.online.grid} show that our proposed online attacks with \NTJAttack~are highly effective for both the criteria: learner is forced to follow the target policy while the attacker's cost is low. In contrast, we can see that the online attacks with \JAttack~lead to high cost for the attacker, i.e., the cumulative cost is linear w.r.t. time as anticipated in Section~\ref{sec.on.problem} (see discussions following Lemma~\ref{lemma.on.known.nontarget}). Second, when comparing \avgmissm~and \avgcost~w.r.t. time $t$ for \NTJAttack~in these two settings, we see that the average values continue to decay for $\gamma=1$, whereas they saturate for $\gamma=0.99$. This is because of the convergence guarantees of the RL agent's learning algorithm: the Q-learning algorithm used for $\gamma=0.99$ has a constant exploration rate whereas the UCRL algorithm used for $\gamma=1$ has no-regret guarantees which in turn leads to $o(1)$  average mismatch and average attack cost  (see Theorems~\ref{theorem.on.regret}~and~\ref{theorem.on.subopt}, and Appendix~\ref{appendix_background}).

\section{Conclusion and Future Work}\label{sec.conclusions}
We studied a security threat to reinforcement learning (RL) where an attacker poisons the environment, thereby forcing the agent into executing a target policy. Our work provides theoretical underpinnings of environment poisoning against RL along several new attack dimensions, including (i) adversarial manipulation of the rewards and transition dynamics jointly, (ii) a general optimization framework for attack against RL agents maximizing rewards in undiscounted or discounted infinite horizon settings, and (iii) analyzing different attack costs for offline planning and online learning settings.

\looseness-1There are several promising directions for future work.
These include expanding the attack models (e.g., manipulating actions or observations of the agent) and broadening the set of attack goals (e.g., under partial specification of target policy). At the same time, relaxing the assumptions on the attacker's knowledge of the underlying MDP could lead to more robust attack strategies. Another interesting future direction would be to make the studied attack models more scalable, e.g., applicable to continuous and large environments. Another interesting topic would be to devise attack strategies against RL agents that use transfer learning approaches, especially in multi-agent RL systems, see \cite{da2019survey}.


While the experimental results demonstrate the effectiveness of the studied attack models, they do not reveal which types of learning algorithms are most vulnerable to the attack strategies studied in the paper. Further experimentation using a diverse set of the state of the art learning algorithms could reveal this, and provide some guidance in designing defensive strategies and novel RL algorithms robust to manipulations.

\section*{Acknowledgements}
Xiaojin Zhu is supported in part by NSF 1545481, 1623605, 1704117, 1836978 and the MADLab AF Center of Excellence FA9550-18-1-0166.



\clearpage
\bibliography{main}

\iftoggle{longversion}{
\clearpage
\onecolumn
\appendix 
{\allowdisplaybreaks
\section{List of Appendices}\label{appendix:table-of-contents}
In this section we provide a brief description of the content provided in the appendices of the paper.
\begin{itemize}
\item Appendix~\ref{appendix_background} gives a few concrete examples of the learning agents considered in this paper. (Section~\ref{sec.preliminaries})
\item Appendix~\ref{appendix.sec.experiments} contains implementation details for different attack strategies used in numerical simulations. (Section~\ref{sec.experiments})
\item Appendix~\ref{appendix.off.general} contains proof of Lemma~\ref{lemma.using_neighbors} and some general results. (Section~\ref{sec.offlineattacks})
\item Appendix~\ref{appendix.off.joint} contains proof of Theorem~\ref{theorem.off.joint} for offline attacks and related discussions. (Section~\ref{sec.offlineattacks})
\item Appendix~\ref{appendix.on} contains proofs for online attacks including 
Lemma~\ref{lemma.on.subopt.missmatch},
Theorem \ref{theorem.on.regret},
and Theorem \ref{theorem.on.subopt}. (Section~\ref{sec.onlineattacks})
\end{itemize}
\section{Examples of Online Learning Agents}
\label{appendix_background}

In this appendix, we provide examples of online learning agents for each of the two settings of interest: average reward optimality criteria with $\gamma = 1$, and discounted reward optimality criteria with $\gamma < 1$.  

\paragraph{Average reward optimality criteria.} For the case of average reward criteria with $\gamma = 1$, we consider a regret-minimization learner. Performance of a regret-minimization learner in MDP $M$ is measured by its \emph{regret} which after $T$ steps is given by $\regret(T, M) = \rho^* \cdot T - \sum_{t=0}^{T-1} r_t$, where $\rho^* := \rho(\pi^*, M)$ is the optimal score. Well-studied algorithms with sublinear regret exist for average reward criteria, e.g., UCRL algorithm~\cite{auer2007logarithmic,jaksch2010near} and algorithms based on posterior sampling method~\cite{agrawal2017optimistic}. More concretely, for the UCRL algorithm, with probability $1-\delta$ we have
$$
\regret(T, M) \le 34 \cdot D|S|\sqrt{|A|T\log\left(\frac{T}{\delta}\right)},
$$
where $D$ is the diameter of MDP~\cite{auer2007logarithmic,jaksch2010near}.


\paragraph{Discounted reward optimality criteria.} For the case of discounted reward criteria with $\gamma < 1$, the type of learners we consider are evaluated based on the number of suboptimal steps they take. An agent is suboptimal at time step $t$ if it takes an an action not used by any near-optimal policy. This is formulated as   $\subopt(T, M, \epsilon') = \sum_{t=0}^{T-1} \ind{a_t \notin \{ \pi(s_t)~|~\rho^\pi \ge \rho^{\pi^*} - \epsilon' \} }$ where $\ind{.}$ denotes the indicator function and $\epsilon'$ measures near-optimality of a policy w.r.t. score $\rho$. Our analysis of attacks is based on $\expct{\subopt(T, M, \epsilon')}$ of the learner for a specific value of $\epsilon'$. Some bounds on this quantity are known for existing algorithms such as classic Q-learning~\cite{even2003learning} and Delayed Q-learning~\cite{strehl2006pac}.

As a concrete example, let us consider the classic Q-learning~\cite{even2003learning}. We can obtain an upper bound on $\expct{\subopt(T, M, \epsilon')}$ of this algorithm based on the results from \cite{even2003learning}. Let $\stdQ_t(s, a)$ denote the $\stdQ$-values estimated by the learner at time step $t$ and let $\pi_t$ be the greedy policy with respect to $\stdQ_t$. These values will converge to the $\stdQ$-values of the optimal policy $\pi^*$ denoted by $\stdQ^*$.\footnote{Note that in the classic algorithm, $\stdQ$-values are not shifted as they are in our definition, and thus, we are using the symbol $\stdQ$ instead of $Q$}

Consider any $\epsilon', \delta > 0$. Then, as shown in \cite{even2003learning}, there exists a number $N(\epsilon' / 2, \delta)$ such that with probability of at least $1 - \delta$ we have $\norm{\stdQ_t - \stdQ^*}_\infty \le \epsilon'/2$ for $t \ge  N(\epsilon' / 2, \delta)$.
%
%
This means that for each state $s$, we have:
\begin{align*}
\stdQ^*(s, \pi_t(s))
 \ge \stdQ_t(s, \pi_t(s)) - \frac{\epsilon'}{2}
\ge \stdQ_t(s, \pi^*(s)) - \frac{\epsilon'}{2}
\ge \stdQ^*(s, \pi^*(s)) - \epsilon',
\end{align*}
where the first and the third inequality holds due to $\norm{\stdQ_t - \stdQ^*}_\infty \le \epsilon'/2$ and the second inequality holds given that $\pi_t$ is the greedy policy with respect to $\stdQ_t$. Furthermore, based on the results in \cite{schulman2015trust}, the following holds for any two policies $\pi$ and $\pi'$:
$$
\rho^\pi - \rho^{\pi'} = 
\sum_{s\in\cS}\mu^{\pi'}(s) \cdot \big(\stdQ^\pi(s, \pi(s)) - \stdQ^\pi(s, \pi'(s))\big).
$$
Since $\mu^{\pi'}(s)$ is a distribution, we conclude that for $t \ge  N(\epsilon' / 2, \delta)$,
$$
\rho^{\pi_t} = \rho^{\pi^*} +
\sum_{s\in\cS}\mu^{\pi_t}(s) \cdot \big(\stdQ^*(s, \pi_t(s)) - \stdQ^*(s, \pi^*(s))\big) \ge \rho^{\pi^*} - \epsilon',
$$
which means that $\pi_t$ is near-optimal with a margin $\epsilon'$.

Considering $\beta$ be the exploration rate (i.e., the probability of taking random action instead of the action from greedy policy), with probability of at least $1 - \delta$ we get
\begin{align*}
\expct{\subopt(T, M, \epsilon')} \le \begin{cases}
T & \quad \mbox{for} \; T < N(\epsilon' / 2, \delta)\\
N(\epsilon' / 2, \delta) + \beta \cdot \big(T - N(\epsilon' / 2, \delta)\big)& \quad \mbox{for} \; T \ge N(\epsilon' / 2, \delta)\\
\end{cases}.
\end{align*}

\section{Numerical Simulations: Implementation Details (Section \ref{sec.experiments})}
\label{appendix.sec.experiments}
%
\looseness-1Here, we provide implementation details for attack strategies. For the reproducibility of experimental
results and facilitating research in this area, the source code of our implementation is publicly available.\footnote{Code: \url{https://machineteaching.mpi-sws.org/files/jmlr2020_rl-policy-teaching_code.zip}.}
We note that optimal solutions for the problems corresponding to \RAttack~and \NTJAttack~can be computed efficiently using standard optimization techniques (also, refer to discussions in Section~\ref{sec.off.efficiency} and Section~\ref{sec.on.problem.refromulation}). Problems corresponding to \JAttack~and \DAttack~are computationally more challenging, and we provide a simple yet effective approach towards finding an approximate solution. 

\paragraph{Implementation details for \DAttack.} To obtain a solution for \DAttack, consider the transitions only attack variant of the problems \eqref{prob.off} and \eqref{prob.on} (i.e., $\widehat{R} := \overline{R}$). Then, we obtain an approximate solution to the problem \eqref{prob.off} by iteratively solving the problem~\eqref{prob.on} as follows:
\begin{itemize}
    \item As a first step, we use a simple heuristic to obtain a pool of transition kernels $\{\widetilde{P}\}$  by perturbations of $\overline{P}$ that increase the score $\rho$ of the target policy $\targetpi$. Here, these transition kernels $\widetilde{P}$ in the pool differ from $\overline{P}$ only for the actions taken by the target policy, i.e., for state action pairs $(s, \targetpi(s)) \ \forall s \in S$.
    \item As the second step, we take each of $\widetilde{P}$ from this pool as an input to the problem \eqref{prob.on} instead of $\overline{P}$, which in turn gives us a corresponding pool of solutions $\{\widehat{P}\}$. Then, we pick a solution from this pool of solutions with the minimal cost.
\end{itemize}

For further details about the transitions only attack strategy (\DAttack), we  refer the reader to the earlier version of the paper~\citep{rakhsha2020policy}.
%

\paragraph{Implementation details for \JAttack.} To obtain a solution for \JAttack, we use the similar idea of approximately solving the problem \eqref{prob.off} by iteratively solving the problem~\eqref{prob.on} as follows:
\begin{itemize}
	\item As a first step, we obtain a solution for \RAttack~and \DAttack~using the above mentioned techniques. Let us denote these solutions to rewards only and transitions only poisoning attacks as $\widehat{R}_{\textnormal{only}}$ and $\widehat{P}_{\textnormal{only}}$ respectively. Note that the \DAttack~attack strategy might be infeasible, and in this case we set $\widehat{P}_{\textnormal{only}} := \overline{P}$.
	\item As a second step,  we use a simple heuristic to obtain a pool of rewards denoted as $\{\widetilde{R}\}$ and a pool of transition kernels denoted as $\{\widetilde{P}\}$. The pool $\{\widetilde{R}\}$ is generated by considering convex combinations of $\widehat{R}_{\textnormal{only}}$ and $\overline{R}$, i.e.,  $\widetilde{R} = (1-\alpha) \cdot \widehat{R}_{\textnormal{only}} + \alpha \cdot \overline{R}$  for $\alpha \in [0,1]$ with a desired level of discretization.
	The pool $\{\widetilde{P}\}$ is generated similarly by considering convex combinations of $\widehat{P}_{\textnormal{only}}$ and $\overline{P}$.
 	\item As the final step, we take all possible pairs $ \langle\widetilde{R}, \widetilde{P} \rangle$ of rewards and transition kernels from these pools as an input to the problem \eqref{prob.on} instead of inputting $ \langle\overline{R}, \overline{P} \rangle$, which in turn gives us a corresponding pool of solutions $\{ \langle\widehat{R}, \widehat{P} \rangle\}$. Then, we pick a solution from this pool of solutions with the minimal cost.
 	%
  %
  %
 %
\end{itemize}
We note that the key common idea for obtaining solutions to \DAttack~and \JAttack~is to approximately solve the problem \eqref{prob.off} by iteratively solving the problem~\eqref{prob.on}. One can replace the specific heuristics to generate the pool $\{\widetilde{P}\}$  for \DAttack~and the pool  $\{ \langle\widetilde{R}, \widetilde{P} \rangle\}$ for \JAttack~with alternate methods.  Furthermore, the run time of solving the problem~\eqref{prob.off} would depend on the number of iterations we invoke the problem~\eqref{prob.on} internally, which in turn provides a simple way to trade-off the run time and attack cost.

\section{Proofs for Offline Attacks: Lemma~\ref{lemma.using_neighbors} (Section
\ref{sec.offlineattacks})}
\label{appendix.off.general}


We prove Lemma~\ref{lemma.using_neighbors} through several intermediate results. The first one is the following results of \cite{even2005experts} and \cite{schulman2015trust}.  
\begin{lemma_new}\label{lemma_qrho_relate}
(Lemma 7 in \cite{even2005experts}, Equation (2) in \cite{schulman2015trust}) For two policies $\pi$ and $\pi'$ we have:
$$
\rho^\pi - \rho^{\pi'} = 
\sum_{s\in\cS}\mu^{\pi'}(s)\big(Q^\pi(s, \pi(s)) - Q^\pi(s, \pi'(s))\big).
$$
\end{lemma_new}{}

We will use the following corollary which is a direct consequence of Lemma \ref{lemma_qrho_relate}.
\begin{corollary_new}
\label{gain_diff_neighbor}
For any policy $\pi$ and its neighbor policy $\neighbor{\pi}{s}{a}$ we have:
$$
\rho^\pi - \rho^{\neighbor{\pi}{s}{a}} = 
\mu^{\neighbor{\pi}{s}{a}}(s)\big(Q^\pi(s, \pi(s)) - Q^\pi(s,a)\big).
$$
\end{corollary_new}{}

 Next, we provide a sufficient condition for a policy $\pi$ to be uniquely optimal. 
 \begin{lemma_new}
\label{single_optimal}
If we have $\rho^\pi \ge \rho^{\neighbor{\pi}{s}{a}} + \epsilon$ for every state $s$ and action $a \ne \pi(s)$, and $\epsilon > 0$, then $\pi$ is the only optimal policy.
\end{lemma_new}
\begin{proof}
Let arbitrary $s\in\cS$ and $a\in\cA$ be such that $a\ne\pi(s)$. Based on corollary \ref{gain_diff_neighbor}, we have
\begin{equation}
\label{has_max_q}
Q^\pi(s, \pi(s)) - Q^\pi(s,a) = \frac{\rho^\pi - \rho^{\neighbor{\pi}{s}{a}}}{\mu^{\neighbor{\pi}{s}{a}}(s)} > 0.
\end{equation}{}
Note that in this paper we are focusing on ergodic MDPs,
thus, we know $\mu^{\neighbor{\pi}{s}{a}}(s) > 0$. Now let $\pi'\ne\pi$ be a policy with $\pi'(s') = a' \ne \pi(s')$. Using (\ref{has_max_q}) we have
\begin{equation*}
\rho^\pi - \rho^{\pi'} = 
\sum_{s\in\cS}\mu^{\pi'}(s)\big(Q^\pi(s, \pi(s)) - Q^\pi(s, \pi'(s))\big) \ge \mu^{\pi'}(s')\big(Q^\pi(s', a') - Q^\pi(s',\pi(s'))\big) > 0
\end{equation*}{}
We again used the fact that MDP is ergodic
  to say $\mu^{\pi'}(s') > 0$.
\end{proof}{}


\textbf{Proof of Lemma~\ref{lemma.using_neighbors}
}
We should show that policy $\pi$ is $\epsilon$-robust optimal \emph{iff} we have $\rho^{\pi} \ge \rho^{\neighbor{\pi}{s}{a}} + \epsilon$ for every state $s$ and action $a \ne \pi(s)$.

\begin{proof}
The necessity of the condition follows directly from the definition of $\epsilon$-robust policies. Let us focus on its sufficiency. 

Consider deterministic policies $\pi$, and denote the Hamming distance between two policies $\pi_1$ and $\pi_2$ by $D_H(\pi_1, \pi_2)$, i.e., $D_H(\pi_1, \pi_2) = \sum_{s \in \cS} \ind{\pi_1(s) \ne \pi_2(s)}$ where $\ind{.}$ denotes the indicator function. 
Assume that the condition of the lemma holds for policy $\pi^*$, i.e., that $\rho^{\pi^*} \ge \rho^{\pi_1} + \epsilon$ for all $\pi_1$ s.t. $D_H(\pi_1, \pi^*) = 1$. 

Lemma ~\ref{single_optimal} implies that $\pi^*$ is uniquely optimal. Now, consider policy $\pi_k$ s.t. $D_H(\pi_k, \pi^*) = k > 1$. Since $\pi^*$ is (uniquely) optimal and the MDP is ergodic
 ($\mu^{\pi^*}(s) > 0$), we have that
\begin{align*}
\rho^{\pi^*} - \rho^{\pi_k} = \sum_{s \in \cS} \mu^{\pi^*}(s) \cdot [Q^{\pi_k}(s, \pi^*(s)) - Q^{\pi_k}(s, \pi_{k}(s))] > 0,
\end{align*}
which implies that there exists $s_k \in \cS$ s.t. $[Q^{\pi_k}(s_k, \pi^*(s_k)) - Q^{\pi_k}(s_k, \pi_{k}(s_k))] > 0$. 
Define policy $\pi_{k-1}$ as
\begin{align*}
\pi_{k-1}(s) = \begin{cases}
\pi^*(s) &\mbox{ if } s = s_k \\
\pi_k(s) & \mbox{ otherwise }
\end{cases}.
\end{align*}
We have that
\begin{align*}
 \rho^{\pi_{k-1}} &= \rho^{\pi_k}  + \sum_{s \in \cS}\mu^{\pi_{k-1}}(s) \cdot [Q^{\pi_k}(s, \pi_{k-1}(s)) - Q^{\pi_k}(s, \pi_{k}(s))]
  \\&=  \rho^{\pi_k}  + \mu^{\pi_{k-1}}(s_k) \cdot [Q^{\pi_k}(s_k, \pi^*(s_k)) - Q^{\pi_k}(s_k, \pi_{k}(s_k))] \ge \rho^{\pi_k}.
\end{align*}
Therefore, by induction, we know that there exists a policy $\pi_1$ such that $\rho^{\pi_k} \le \rho^{\pi_1}$ and $D_H(\pi_1, \pi^*) = 1$. Utilizing our initial assumption, we obtain that $\rho^{\pi^*} \ge \rho^\pi + \epsilon$ for all $\pi \ne \pi^*$, which proves that $\pi^*$ is $\epsilon$-robust optimal if the condition of the lemma
 holds. 
\end{proof}

\section{Proofs for Offline Attacks: Proof of Theorem \ref{theorem.off.joint}}
\label{appendix.off.joint}
We break the proof of Theorem~\ref{theorem.off.joint} into two parts: In Appendix \ref{appendix.off.lower}, we prove the lower bound in the theorem, and in Appendix \ref{appendix.off.upper} we prove the upper bound.
In Appendix \ref{appendix.off.discussion}, we discuss the choice of $\epsP$ in our optimization problems.

\subsection{Proofs for the Lower Bound}
\label{appendix.off.lower}
To prove the lower bound in Theorem~\ref{theorem.off.joint}, we 
will use a proof technique that is similar to the one presented in \cite{ma2019policy}, but adapted to our setting.  

First, let us define operator $F$ as
\begin{align}\label{contraction_operator}
F(Q, R, \rho, P, \pi, \gamma)(s,a) = R(s, a) - \rho + \gamma\sum_{s' \in \cS} P(s,a,s') V^\pi(s'),
\end{align}
or in vector notation
\begin{align*}
F(Q, R, \rho, P, \pi, \gamma) = R - \rho \cdot \mathbf 1 + \gamma P \cdot V^\pi,
\end{align*}
where $V^\pi(s') = Q(s',\pi(s'))$ (and $\pi$ is a deterministic policy). Furthermore, we defined the span of $X$ as $sp(X) = \max_i X(i) - \min_i X(i)$ - as argued in \cite{Puterman1994}, $sp$ is a seminorm. 

\begin{lemma_new}\label{lemma_contraction}
	The following holds: 
	\begin{align*}
	sp(F(Q_1, R, \rho, P, \pi, \gamma) - F(Q_2, R, \rho, P, \pi, \gamma)) \le \gamma(1-\alpha) \cdot sp(Q_1 - Q_2),
	\end{align*}
	where 
	\begin{align*}
	\alpha = \min_{s,a,s',a'} \sum_{x \in \cS}\min \{P(s,a,x), P(s',a',x) \}.
	\end{align*}
\end{lemma_new}
\begin{proof}
	We have that 
	\begin{align*}
	sp(F(Q_1, R, \rho, P, \pi, \gamma) - F(Q_2, R, \rho, P, \pi, \gamma)) = \gamma \cdot sp(P \cdot (V_{1}^\pi - V_{2}^\pi)). 
	\end{align*}
	Following the proof of Proposition 6.6.1 in \cite{Puterman1994}, we obtain that for $b(x,s,a,s',a') = \min \{P(s,a,x), P(s',a',x) \}$
	\begin{align*}
	&\sum_{x \in \cS} P(s,a,x) \cdot  (V_{1}^\pi(x) - V_{2}^\pi(x)) -  \sum_{x' \in \cS} P(s', a',x)  \cdot (V_{1}^\pi(x) - V_{2}^\pi(x))\\
	&= \sum_{x \in \cS} (P(s,a,x) - b(x,s,a,s',a')) \cdot  (V_{1}^\pi(x) - V_{2}^\pi(x)) \\
	&-  \sum_{x \in \cS} (P(s', a',x)  - b(x,s,a,s',a')))  \cdot (V_{1}^\pi(x) - V_{2}^\pi(x))\\
	&\le \sum_{x \in \cS} (P(s,a,x) - b(x,s,a,s',a')) \cdot \max_{x'} (V_{1}^\pi(x') - V_{2}^\pi(x')) \\
	&-  \sum_{x \in \cS} (P(s', a',x)  - b(x,s,a,s',a')))  \cdot \min_{x'} (V_{1}^\pi(x') - V_{2}^\pi(x'))\\
	&= (1 - \sum_{x\in \cS} b(x,s,a,s',a')) \cdot sp(V_{1}^\pi - V_{2}^\pi) \le (1-\alpha) \cdot sp(V_{1}^\pi - V_{2}^\pi)
	\end{align*}
	Therefore
	\begin{align*}
	sp(F(Q_1, R, \rho, P, \pi, \gamma) - F(Q_2, R, \rho, P, \pi, \gamma)) = \gamma \cdot sp(P \cdot (V_{1}^\pi - V_{2}^\pi)) \le \gamma(1 - \alpha) \cdot sp(V_{1}^\pi - V_{2}^\pi).
	\end{align*}
	Now, notice that for $s_\textnormal{max} = \arg \max_s [V_{1}^\pi(s) - V_{2}^\pi(s)]$ and $s_\textnormal{min} = \arg \min_s [V_{1}^\pi(s) - V_{2}^\pi(s)]$ we have that
	\begin{align*}
	V_{1}^\pi(s_\textnormal{max}) - V_{2}^\pi(s_\textnormal{max}) &= Q_1(s_\textnormal{max},\pi(s_\textnormal{max})) - Q_2(s_\textnormal{max},\pi(s_\textnormal{max})) \le  \max_{s,a} [Q_1(s,a) - Q_2(s,a)] \\
	V_{1}^\pi(s_\textnormal{min}) - V_{2}^\pi(s_\textnormal{min}) &= Q_1(s_\textnormal{min},\pi(s_\textnormal{min})) - Q_2(s_\textnormal{min},\pi(s_\textnormal{min})) \ge   \min_{s, a} [Q_1(s,a) - Q_2(s,a)]
	\end{align*}
	Therefore 
	$sp(V_{1}^\pi - V_{2}^\pi) \le sp(Q_1 - Q_2)$,
	which implies that
	\begin{align*}
	sp(F(Q_1, R, \rho, P, \pi, \gamma) - F(Q_2, R, \rho, P, \pi, \gamma)) \le \gamma(1-\alpha) \cdot sp(Q_1 - Q_2)
	\end{align*}
\end{proof}

To obtain the statement of the theorem, we will need to relate $sp(Q_1 - Q_2)$ to difference between $R_1, P_1$ and $R_2, P_2$. The following lemma provides this relation.

\begin{lemma_new}\label{lemma_qr_relation}
	Let $Q_1^\pi$ and $V_1^\pi$ denote Q and V values of policy $\pi$ in MDP $M_1 = (\cS, \cA, R_1, P_1, \gamma)$ and $Q_2^\pi$ denote Q values of policy $\pi$ in MDP $M_2 = (\cS, \cA, R_2, P_2, \gamma)$. The following holds:
	\begin{align*}
	\norm{R_1 - R_2}_{\infty} + \gamma \cdot \norm{P_1 - P_2}_{\infty} \cdot \norm{V_1^\pi}_{\infty} \ge \frac{1- \gamma + \gamma \cdot \alpha_2}{2} \cdot sp(Q_1^\pi - Q_2^\pi).
	\end{align*}
	where
	\begin{align*}
	\alpha_2 = \min_{s,a,s',a'} \sum_{x \in \cS}\min \{P_2(s,a,x), P_2(s',a',x) \}.
	\end{align*}
	
\end{lemma_new}
\begin{proof}
	Notice that $Q_1^\pi$ and $Q_2^\pi$ satisfy 
	\begin{align*}
	Q_1^\pi(s,a) &= F(Q_1^\pi, R_1, \rho_1^\pi, P_1, \pi, \gamma)\\
	Q_2^\pi(s,a) &= F(Q_2^\pi, R_2, \rho_2^\pi, P_2, \pi, \gamma),
	\end{align*}
	where $\rho_1^\pi$ and $\rho_2^\pi$ respectively denote the average rewards of policy $\pi$ in $M_1$ and $M_2$. We obtain
	\begin{align*}
	sp(Q_1^\pi - Q_2^\pi) &= sp(F(Q_1^\pi, R_1, \rho_1^\pi, P_1, \pi, \gamma) - F(Q_2^\pi, R_2, \rho_2^\pi, P_2, \pi, \gamma))\\
	&= sp(F(Q_1^\pi, R_1, \rho_1^\pi, P_1, \pi, \gamma) 
	-F(Q_1^\pi, R_2, \rho_2^\pi, P_1, \pi, \gamma)
	\\&+F(Q_1^\pi, R_2, \rho_2^\pi, P_1, \pi, \gamma)
	-F(Q_1^\pi, R_2, \rho_2^\pi, P_2, \pi, \gamma)
	\\&+F(Q_1^\pi, R_2, \rho_2^\pi, P_2, \pi, \gamma)
	-F(Q_2^\pi, R_2, \rho_2^\pi, P_2, \pi, \gamma))\\ 
	&\le sp(F(Q_1^\pi, R_1, \rho_1^\pi, P_1, \pi, \gamma) 
	-F(Q_1^\pi, R_2, \rho_2^\pi, P_1, \pi, \gamma))
	\\&+sp(F(Q_1^\pi, R_2, \rho_2^\pi, P_1, \pi, \gamma)
	-F(Q_1^\pi, R_2, \rho_2^\pi, P_2, \pi, \gamma))
	\\&+sp(F(Q_1^\pi, R_2, \rho_2^\pi, P_2, \pi, \gamma)
	-F(Q_2^\pi, R_2, \rho_2^\pi, P_2, \pi, \gamma))\\
	&\le sp(R_1 - \rho_1^\pi \cdot \mathbf 1 - R_2 + \rho_2^\pi \cdot \mathbf 1)
	+ \gamma \cdot sp((P_1-P_2) \cdot V_1^\pi) + \gamma(1-\alpha_2) \cdot sp(Q_1^\pi - Q_2^\pi)
	\end{align*}
	where the last inequality is due to Lemma \ref{lemma_contraction} (i.e., $sp(F(Q_1^\pi, R_2, \rho_2^\pi, P_2, \pi, \gamma) - F(Q_2^\pi, R_2, \rho_2^\pi, P_2, \pi, \gamma)) \le \gamma(1-\alpha_2) \cdot sp(Q_1^\pi - Q_2^\pi)$). 
	Due to the properties of $sp$, we have
	\begin{align*}
	sp(R_1 - \rho_1^\pi \cdot \mathbf 1 - R_2 + \rho_2^\pi \cdot \mathbf 1) = sp(R_1 - R_2) \le 2\cdot \norm{R_1 - R_2}_{\infty}.
	\end{align*}
	For the second term, we have  
	\begin{align*}
	sp((P_1-P_2) \cdot V_1^\pi) &\le 2\cdot \norm{(P_1-P_2) \cdot V_1^\pi}_{\infty} \\
	&=2 \cdot \max_{s,a}|\sum_{s'}( P_1(s,a,s') -  P_2(s,a,s')) \cdot V_1^\pi(s')|.
	\end{align*}	
	To bound the right-hand side in the above equation, we note that
	\begin{align*}
		&\sum_{s'}( P_1(s,a,s') -  P_2(s,a,s')) \cdot V_1^\pi(s') \\
		&\le \bigg(
		\sum_{s': P_1(s,a,s') \ge  P_2(s,a,s')}( P_1(s,a,s') -  P_2(s,a,s')) 
		\bigg)\cdot \max_{s'} V_1^\pi(s') \\
		&+ \bigg(
		\sum_{s': P_1(s,a,s') <  P_2(s,a,s')}( P_1(s,a,s') -  P_2(s,a,s')) 
		\bigg)\cdot \min_{s'} V_1^\pi(s') \\
		&= \frac{1}{2} \bigg(
		\sum_{s'}|P_1(s,a,s') -  P_2(s,a,s')| 
		\bigg)\cdot sp(V_1^\pi),
	\end{align*}
	and similarly
	\begin{align*}
	&\sum_{s'}( P_1(s,a,s') -  P_2(s,a,s')) \cdot V_1^\pi(s') \\
	&\ge \bigg(
	\sum_{s': P_1(s,a,s') \ge  P_2(s,a,s')}( P_1(s,a,s') -  P_2(s,a,s')) 
	\bigg)\cdot \min_{s'} V_1^\pi(s') \\
	&+ \bigg(
	\sum_{s': P_1(s,a,s') <  P_2(s,a,s')}( P_1(s,a,s') -  P_2(s,a,s')) 
	\bigg)\cdot \max_{s'} V_1^\pi(s') \\
	&= -\frac{1}{2} \bigg(
	\sum_{s'}|P_1(s,a,s') -  P_2(s,a,s')| 
	\bigg)\cdot sp(V_1^\pi).
	\end{align*}
	The above two bounds give us the following:
	\begin{align*}
		|\sum_{s'}( P_1(s,a,s') -  P_2(s,a,s')) \cdot V_1^\pi(s')| \le 
		\frac{1}{2} \bigg(
		\sum_{s'}|P_1(s,a,s') -  P_2(s,a,s')| 
		\bigg)\cdot sp(V_1^\pi).
	\end{align*}
	Now we can bound the second term as
	\begin{align*}
	sp((P_1-P_2) \cdot V_1^\pi) &\le 2\cdot \norm{(P_1-P_2) \cdot V_1^\pi}_{\infty} \\
	&=2 \cdot \max_{s,a}|\sum_{s'}( P_1(s,a,s') -  P_2(s,a,s')) \cdot V_1^\pi(s')|\\
	&= sp(V_1^\pi) \cdot \max_{s,a} \sum_{s'} |P_1(s,a,s') -  P_2(s,a,s')| \\
	&= sp(V_1^\pi) \cdot \norm{P_1 - P_2}_\infty,
	\end{align*}	
	where $\norm{P_1 - P_2}_{\infty} = \max_{s,a} \sum_{s'}|P_1(s,a,s') - P_2(s,a,s')|$.
	Putting this together with the upper bound on $sp(Q_1^\pi - Q_2^\pi)$, we obtain
	\begin{align*}
	2 \cdot \norm{R_1 - R_2}_{\infty} + \gamma \cdot \norm{P_1 - P_2}_{\infty} \cdot sp(V_1^\pi)\ge (1 - \gamma + \gamma \alpha_2) \cdot sp(Q_1^\pi - Q_2^\pi),
	\end{align*}
	which proves the claim. 
\end{proof}{}
\vspace{-2mm}
We are now ready to prove the lower bound in Theorem \ref{theorem.off.joint}. We can write
\begin{align*}
\cost(\widehat{M}, \overline{M}) =& \bigg[\sum_{s,a} \Big(
\costr \cdot \big|\widehat{R}(s,a) - \overline{R}(s,a)\big|
+ 
\costp \cdot \sum_{s'}\big|\widehat{P}(s,a,s') - \overline{P}(s,a,s')\big|
\Big)^p
\bigg]^{1/p}\\
\ge& \max_{s, a} \Big( \costr \cdot \big|\widehat{R}(s,a) - \overline{R}(s,a)\big|
+ 
\costp \cdot \sum_{s'}\big|\widehat{P}(s,a,s') - \overline{P}(s,a,s')\big| \Big)\\
\ge& \max \left(
\costr \norm{\widehat R - \overline R}_\infty, \costp\norm{\widehat P - \overline P}_\infty
\right)\\
\ge& \frac{2\costr^{-1}}{2\costr^{-1} + \gamma \costp^{-1} sp(\overline V^{\targetpi})} \costr \norm{\widehat R - \overline R}_\infty
+
\frac{\gamma \costp^{-1} sp(\overline V^{\targetpi})}{2\costr^{-1} + \gamma \costp^{-1} sp(\overline V^{\targetpi})} \costp\norm{\widehat P - \overline P}_\infty\\
=&
\frac{1}{2\costr^{-1} + \gamma \costp^{-1} sp(\overline V^{\targetpi})}
\left(
2\norm{\widehat R - \overline  R}_\infty + 
\gamma sp(\overline V^{\targetpi}) \norm{\widehat P - \overline P}_\infty
\right)\\
\ge& 
\frac{1 - \gamma + \gamma \widehat{\alpha}}
{2\costr^{-1} + \gamma \costp^{-1} sp(\overline V^{\targetpi})}
sp(\overline{Q}^{\targetpi} - \widehat{Q}^{\targetpi}),
\end{align*}
where $\norm{P_1 - P_2}_{\infty} = \max_{s,a} \sum_{s'}|P_1(s,a,s') - P_2(s,a,s')|$, and we used Lemma~\ref{lemma_qr_relation} in the last inequality. Factor $\widehat{\alpha}$ can be bounded as follows:
\begin{align*}
\widehat{\alpha} &= \min_{s,a,s',a'} \sum_{x} \min \{ \widehat{P}(s,a,x) , \widehat{P}(s',a',x) \} \\
&\ge \min_{s,a,s',a'} \sum_{x} \min \{ \delta \cdot \overline{P}(s,a,x) , \delta \cdot  \overline{P}(s',a',x) \} \\
&= \delta \cdot \min_{s,a,s',a'} \sum_{x} \min \{ \overline{P}(s,a,x) , \cdot  \overline{P}(s',a',x) \} \\
&= \delta \cdot \overline{\alpha}.
\end{align*}
It only remains to bound $sp(\overline{Q}^{\targetpi} - \widehat{Q}^{\targetpi})$. We show that
\begin{align*}
sp(\overline{Q}^{\targetpi} - \widehat{Q}^{\targetpi}) \ge \norm{\overline{\chi}^{\targetpi}_0}_\infty.
\end{align*}

Let $s'$ and $a'$ be a state action pair that satisfy: $s',a' = \arg\max_{s,a} \overline{\chi}_0^{\targetpi}( s, a)$.
Let's consider the case when $\overline{\chi}_0^{\targetpi}(s', a') > 0$.
We have
\begin{align*}
sp(\overline{Q}^{\targetpi} - \widehat{Q}^{\targetpi}) &= sp(\widehat{Q}^{\targetpi} - \overline{Q}^{\targetpi}) = \max_{s,a} [\widehat{Q}^{\targetpi} - \overline{Q}^{\targetpi}]
- \min_{s,a}[\widehat{Q}^{\targetpi} - \overline{Q}^{\targetpi} ] \\
&\ge 
\widehat{Q}^{\targetpi}(s', \targetpi(s')) - \overline{Q}^{\targetpi}(s', \targetpi(s')) - (\widehat{Q}^{\targetpi}(s', a') - \overline{Q}^{\targetpi}(s', a')) \\
&=
(\widehat{Q}^{\targetpi}(s', \targetpi(s')) - \widehat{Q}^{\targetpi}(s', a')) +  
(\overline{Q}^{\targetpi}(s', a') - \overline{Q}^{\targetpi}(s', \targetpi(s')) \\
&\ge \frac{\epsilon}{\widehat{\mu}^{\neighbor{\targetpi}{s'}{a'}}(s')} + (\overline{Q}^{\targetpi}(s', a') - \overline{Q}^{\targetpi}(s', \targetpi(s')) \\
&\ge \epsilon + \frac{\overline{\rho}^{\neighbor{\targetpi}{s'}{a'}} - \overline{\rho}^{\targetpi}}{\overline{\mu}^{\neighbor{\targetpi}{s'}{a'}}(s')} > \overline{\chi}^{\targetpi}_0(s', a'),
\end{align*}
where we used the fact that $\widehat{Q}^{\targetpi}(s', \targetpi(s')) \ge \widehat{Q}^{\targetpi}(s', a') +  \frac{\epsilon}{\widehat{\mu}^{\neighbor{\targetpi}{s'}{a'}}(s')}$ (because $\targetpi$ is $\epsilon$-robust optimal in the modified MDP) and Lemma \ref{lemma_qrho_relate} (Corollary \ref{gain_diff_neighbor}) to relate Q values to scores $\rho$.
When $\overline{\chi}_0^{\targetpi}(s', a') = 0$, we know that $sp(\overline{Q}^{\targetpi} - \widehat{Q}^{\targetpi}) \ge 0$ due to the properties of $sp$. Therefore, $sp(\overline{Q}^{\targetpi} - \widehat{Q}^{\targetpi}) = sp(\widehat{Q}^{\targetpi} - \overline{Q}^{\targetpi}) \ge \norm{\overline{\chi}^{\targetpi}_0(s', a')}_{\infty}$. 
Putting this together with the previous expression, we obtain the claim:
\begin{align*}
\cost(\widehat{M}, \overline{M}) \ge \frac{1 - \gamma + \gamma \delta \overline{\alpha}}
{2\costr^{-1} + \gamma \costp^{-1} sp(\overline V^{\targetpi})}
\norm{\overline{\chi}^{\targetpi}_0}_\infty.
\end{align*}

\subsection{Proofs for the Upper Bound}
\label{appendix.off.upper}
Here, we prove the upper bound in Theorem~\ref{theorem.off.joint}. We first prove a lemma that we need for our proof.
\begin{lemma_new}
	\label{lemma.mu.equality.bound}
	Let $\pi$ be a deterministic policy and $P$ be a transition kernel such that $P(s, \pi(s), s') = \overline{P}(s, \pi(s), s')$ for every $s, s'$. For any $s, a$, If $\mu^{\neighbor{\pi}{s}{a}}$ is the state distribution of $\neighbor{\pi}{s}{a}$ under $P$ and initial state distribution $d_0$, we have 
	\begin{align*}
	\mu^{\neighbor{\pi}{s}{a}}(s) =  \frac
	{1 - (1 - \gamma)\sum_{s'}d_0(s')\overline{T}^{\pi}(s', s)}
	{1 + \gamma \sum_{s'}P(s, a, s')\overline{T}^{\pi}(s', s)}
	\ge
	\frac
	{1 - (1 - \gamma)\cdot \overline{D}^\pi}
	{1 + \gamma \cdot \overline{D}^\pi}.
	\end{align*}{}
\end{lemma_new}
\begin{proof}
The inequality is trivial due to the fact that by definition
$\overline{D}^{\pi} \ge \overline{T}^{\pi}(s, s')$ for every $s, s'$. Thus, it suffices to prove the equality.

When $\gamma = 1$, we have $\frac{1}{\mu^{\neighbor{\targetpi}{s}{a}}(s)} = \expct{L^{\neighbor{\targetpi}{s}{a}}(s, s)}$ (see Theorem~1.21 in \cite{durrett1999essentials}), where $L^{\pi}(s, s')$ is the number of steps to reach $s'$ starting from $s$ in the induced Markov chain by $\pi$ in $P$. Note that as $\gamma = 1$, the discounted reach times on $P$ are $T^\pi(s, s') = \expct{L^{\pi}(s, s')}$ for $s \ne s'$. We have
\begin{align*}
\frac{1}{\mu^{\neighbor{\targetpi}{s}{a}}(s)}
&=  \expct{L^{\neighbor{\targetpi}{s}{a}}(s, s)}\\
&=  1 + \sum_{s'} P(s, a, s') \cdot T^{\neighbor{\targetpi}{s}{a}}(s', s)\\
&=1 +  \sum_{s'} P(s, a, s') \cdot T^{\targetpi}(s', s)\\
&=1 + \sum_{s'} P(s, a, s') \cdot \overline{T}^{\targetpi}(s', s).
\end{align*}
Note that we have used the fact that transitions of $\targetpi$ are not changed, and therefore $\overline{T}^{\targetpi} = T^{\targetpi}$.

Now consider the case $\gamma < 1$. For $i \ge 1$, let $t_i(s)$ be the random variable denoting the step number when state $s$ is visited for the $i$-th time. More formally
\begin{gather*}
t_1(s) = \min \{j\ge 0 : s_j = s\},\\
t_i(s) = \min \{j > t_{i - 1}(s) : s_j = s \}.
\end{gather*}
Using the definition of $\mu^{\neighbor{\pi}{s}{a}}(s)$ and independence of $t_i(s) - t_1(s)$ and $t_1(s)$, we obtain 
\begin{align*}
\mu^{\neighbor{\pi}{s}{a}}(s) 
&= (1 - \gamma) \sum_{t = 0}^{\infty} \gamma^t\Pr{s_t = s | s_0 \sim d_0, \neighbor{\pi}{s}{a}}\\
\\
&= (1 - \gamma)
\expct{\sum_{i = 1}^\infty \gamma^{t_i(s)}|s_0 \sim d_0, \neighbor{\pi}{s}{a}}\\
&= (1 - \gamma)\expct{
	\gamma^{t_1(s)}\Big(1 + 
	\sum_{i = 2}^\infty \gamma^{t_i(s) - t_1(s)}\Big)|s_0 \sim d_0, \neighbor{\pi}{s}{a}}\\
&= (1 - \gamma)\expct{
	\gamma^{t_1(s)} |s_0 \sim d_0, \neighbor{\pi}{s}{a}} \expct{\Big(1 + 
	\sum_{i = 2}^\infty \gamma^{t_i(s) - t_1(s)}\Big)|s_0 \sim d_0, \neighbor{\pi}{s}{a}}\\
&= (1 - \gamma)\expct{
	\gamma^{t_1(s)}|s_0 \sim d_0, \neighbor{\pi}{s}{a}}\bigg(1 + \gamma\expct{
	\sum_{i = 1}^\infty \gamma^{t_i(s)} | s_0 \sim P(s, a, .), \neighbor{\pi}{s}{a}}\bigg)\\
&= (1 - \gamma)\expct{
	\gamma^{t_1(s)}|s_0 \sim d_0, \neighbor{\pi}{s}{a}}\bigg(1 + \frac{\gamma\mu^{\neighbor{\pi}{s}{a}}_{P(s, a, .)}(s)}{1 - \gamma}\bigg)\\
&= \expct{
	\gamma^{t_1(s)}|s_0 \sim d_0, \neighbor{\pi}{s}{a}}\Big(1 - \gamma + \gamma\mu^{\neighbor{\pi}{s}{a}}_{P(s, a, .)}(s)\Big).
\end{align*}{}
Here, $\mu^{\neighbor{\pi}{s}{a}}_{P(s, a, .)}$ is the state distribution of $\neighbor{\pi}{s}{a}$ under $P$ when the initial state
 distribution is $P(s, a, .)$ instead of $d_0$. 
For an arbitrary policy $\pi'$ define $X^{\pi'}(s, s')$ as
\begin{align*}
X^{\pi'}(s, s') = \expct{\gamma^{t_1(s')}|s_0 = s, \pi'}.
\end{align*}{}
Using this, we can write the last equation as
\begin{align}
\label{eq_by_P_mu}
\mu^{\neighbor{\pi}{s}{a}}(s)
&= \bigg(\sum_{s'}d_0(s')X^{\neighbor{\pi}{s}{a}}(s', s)
\bigg)\big(1 - \gamma + \gamma \mu^{\neighbor{\pi}{s}{a}}_{P(s, a, .)}(s)\big).
\end{align}{}
Similarly
\begin{align*}
&\mu^{\neighbor{\pi}{s}{a}}_{P(s, a, .)}(s)
= \bigg(\sum_{s'}P(s, a, s')X^{\neighbor{\pi}{s}{a}}(s', s)
\bigg)\big(1 - \gamma + \gamma \mu^{\neighbor{\pi}{s}{a}}_{P(s, a, .)}(s)\big)\\
\Rightarrow &
\mu^{\neighbor{\pi}{s}{a}}_{P(s, a, .)}(s) = 
\frac{(1 - \gamma)\bigg(\sum_{s'}P(s, a, s')X^{\neighbor{\pi}{s}{a}}(s', s)
	\bigg)}{1 -  \gamma \sum_{s'}P(s, a, s')X^{\neighbor{\pi}{s}{a}}(s', s)}.
\end{align*}{}
Plugging this into (\ref{eq_by_P_mu}) we get
\begin{align*}
\mu^{\neighbor{\pi}{s}{a}}(s)
&=
\bigg(\sum_{s'}d_0(s')X^{\neighbor{\pi}{s}{a}}(s', s)\bigg)\cdot \frac{1 - \gamma}
{1 - \gamma \sum_{s'}P(s, a, s')X^{\neighbor{\pi}{s}{a}}(s', s)}.
\end{align*}{}
Finally, note that $X^{\neighbor{\pi}{s}{a}}(s', s) = X^{\pi}(s', s)$ as $\neighbor{\pi}{s}{a}$ and $\pi$ only differ in $s$ which is does not affect the time to visit $s$ for the first time. Moreover, we have $X^{\pi}(s', s) = 1 - \overline T^{\pi}(s', s)(1 - \gamma)$ because for $s \ne s'$ one can write
\begin{align*}
X^{\pi}(s', s) &= \expct{\gamma^{t_1(s')}|s_0 = s, \pi}\\
&=\expct{\gamma^{\overline{L}^{\pi}(s, s')}}\\
&=\expct{1 - \frac{1 -\gamma^{\overline{L}^{\pi}(s, s')}}{1 - \gamma} \cdot (1 - \gamma)}\\
&= 1 - \overline T^{\pi}(s', s)(1 - \gamma).
\end{align*}
and for $s' = s$, $X^{\pi}(s, s) = 1$ and $\overline{T}^{\pi}(s, s) = 0$.
We get
\begin{align*}
\mu^{\neighbor{\pi}{s}{a}}(s)
&=\frac{(1 - \gamma)\sum_{s'}d_0(s') X^{\pi}(s', s)}
{1 - \gamma \sum_{s'}P(s, a, s') X^{\pi}(s', s)}\\
&= \frac{(1 - \gamma)\sum_{s'}d_0(s') (1 -  (1 - \gamma)\overline T^{\pi}(s', s))}
{1 - \gamma \sum_{s'}P(s, a, s') (1 -  (1 - \gamma)\overline T^{\pi}(s', s))}\\
&= \frac{(1 - \gamma)(1 - \sum_{s'}d_0(s') (1 - \gamma)\overline T^{\pi}(s', s))}
{1 - \gamma( 1 -  \sum_{s'}P(s, a, s') (1 - \gamma)\overline T^{\pi}(s', s))}\\
&= \frac{1 - (1 - \gamma)\sum_{s'}d_0(s') \overline T^{\pi}(s', s)}
{1 + \gamma( \sum_{s'}P(s, a, s') \overline T^{\pi}(s', s))},
\end{align*}{}
which completes the proof.
\end{proof}

Now we can prove the upper bound in Theorem~\ref{theorem.off.joint}. First, note that $k(s, a) < |S|$ due to the condition $\gamma \costr (\overline V^{\targetpi}(s_{k(s, a)}) - \overline V^{\targetpi}(s_{|S|})) > 2\costp$ and $\costp \ge 0$. Consider the following solution:
\begin{gather}
\widehat{P}(s, a, s_i) =
\begin{cases}
\delta \overline{P}(s, a, s_i) & \mbox{if} \; a \ne \targetpi(s) \; \mbox{and}\; i \le k(s, a)\\
\overline{P}(s, a, s_i) + \frac{1}{2} \overline G_{k(s, a)}(s, a)   & \mbox{if} \; a \ne \targetpi(s) \; \mbox{and}\; i = |S|\\
\overline{P}(s, a, s_i) & \mbox{otherwise}\\
\end{cases},\\
\widehat{R}(s, a) = 
\begin{cases}
\overline{R}(s, a) - \overline{\chi}^{\targetpi}_{\overline \beta(s, a)}(s, a) + \overline F_{k(s, a)}(s, a) 
& \mbox{if} \; a \ne \targetpi(s)\\
\overline{R}(s, a)
& \mbox{if} \; a = \targetpi(s)\\
\end{cases}.
\end{gather}

Let $\widehat{Q}$ and $\widehat{V}$ denote $Q$-values and $V$-values in $\widehat{M}$. From Lemma~\ref{lemma.using_neighbors} and Corollary~\ref{gain_diff_neighbor} it suffices to prove for state $s$ and action $a \ne \targetpi(s)$
\begin{align*}
\widehat{V}^{\targetpi}(s) - \widehat{Q}^{\targetpi}(s, a) \ge \frac{\epsilon}{\widehat{\mu}^{\neighbor{\targetpi}{s}{a}}(s)}.
\end{align*}

Note that rewards and transitions used by the target policy are not changed so we have 
$\widehat{V}^{\targetpi}$ = $\overline{V}^{\targetpi}$ and $\widehat{\rho}^{\targetpi} = \overline{\rho}^{\targetpi}$. One can write 
\begin{align*}
\widehat{Q}^{\targetpi}(s, a) =&
\widehat{R}(s, a) - \overline{\rho}^{\targetpi} + \gamma\sum_{i = 1}^{|S|}\widehat{P}(s, a, s_i)\overline{V}^{\targetpi}(s_i)\\
=& \overline{R}(s, a)  - \overline{\chi}^{\targetpi}_{\overline \beta(s, a)}(s, a) + \overline F_{k(s, a)}(s, a)  - \overline{\rho}^{\targetpi} + \gamma\sum_{i = 1}^{|S|}\overline{P}(s, a, s_i)\overline{V}^{\targetpi}(s_i)  \\
&+ \frac{\gamma}{2} \overline G_{k(s, a)}(s, a) \cdot \overline{V}^{\targetpi}(s_{|S|})
- \gamma \cdot (1 - \delta)\sum_{i = 1}^{k(s, a)}\overline{P}(s, a, s_i)\overline{V}^{\targetpi}(s_i)\\
=& \overline{Q}^{\targetpi}(s, a)  - \overline{\chi}^{\targetpi}_{\overline \beta(s, a)}(s, a) + 
\gamma \sum_{i = 1}^{k(s, a)} (1 - \delta)\overline P(s, a, s_i)(\overline V^{\targetpi}(s_i) - 
\overline V^{\targetpi}(s_{|S|}))
\\
&+ \gamma \cdot \Big(\sum_{i = 1}^{k(s, a)}(1 - \delta)\overline{P}(s, a, s_i)\Big) \cdot \overline{V}^{\targetpi}(s_{|S|})
- \gamma \cdot (1 - \delta)\sum_{i = 1}^{k(s, a)}\overline{P}(s, a, s_i)\overline{V}^{\targetpi}(s_i)\\
=& \overline{Q}^{\targetpi}(s, a)  - \overline{\chi}^{\targetpi}_{\overline \beta(s, a)}(s, a) \\
\le& \overline{Q}^{\targetpi}(s, a)  - 
\frac{
	\overline \rho^{\neighbor{\pi}{s}{a}} - \overline \rho^{\pi} + \overline \beta(s,a)
}{
	\overline \mu^{\neighbor{\pi}{s}{a}}(s)
}\\
=&  \overline{V}^{\targetpi}(s)
- \frac{\overline \beta(s,a)}
{\overline \mu^{\neighbor{\pi}{s}{a}}(s)},
\end{align*}
where in the last equality we used Corollary~\ref{gain_diff_neighbor}. By definition 
\begin{align*}
\overline \beta(s, a) =
\epsilon
\cdot
\overline \mu^{\neighbor{\targetpi}{s}{a}}(s)
\cdot
\frac
{1 + \gamma \cdot \overline{D}^{\targetpi}}
{1 - (1 - \gamma) \cdot \overline{D}^{\targetpi}}.
\end{align*}
Thus, from Lemma~\ref{lemma.mu.equality.bound} we can see that 
\begin{align*}
\frac{\overline \beta(s,a)}
{\overline \mu^{\neighbor{\pi}{s}{a}}(s)} \ge 
\frac{\epsilon}{\widehat \mu^{\neighbor{\pi}{s}{a}}(s)}.
\end{align*}
Combining this with the last expression, we obtain 
\begin{align*}
\overline{V}^{\targetpi}(s) - \widehat{Q}^{\targetpi}(s, a) \ge \frac{\epsilon}{\widehat{\mu}^{\neighbor{\targetpi}{s}{a}}(s)}.
\end{align*}
Thus, this is a solution for the problem. It only remains to find its cost. We can write
\begin{align*}
&\cost(\widehat{M}, \overline{M}) = \bigg (\sum_{s,a} \Big(
\costr \cdot \big|\widehat{R}(s,a) - \overline{R}(s,a)\big|
+ 
\costp \cdot \sum_{s'}\big|\widehat{P}(s,a,s') - \overline{P}(s,a,s')\big|
\Big)^p
\bigg)^{1/p}\\
\le& \bigg(\sum_{s,a} \Big(
\costr \cdot \big(\overline{\chi}^{\targetpi}_{\overline \beta(s, a)}(s, a) - \overline F_{k(s, a)}(s, a) \big)
+ 
\costp \cdot \big(
\sum_{i=1}^{k(s,a)}(1 - \delta)\overline{P}(s, a, s_i) + \frac{1}{2} \overline{G}_{k(s,a)}(s, a)
\big)
\Big)^p
\bigg)^{1/p}\\
=& \bigg(\sum_{s,a} \Big(
\costr \cdot \big(\overline{\chi}^{\targetpi}_{\overline \beta(s, a)}(s, a) - \overline F_{k(s, a)}(s, a) \big)
+ 
\costp \cdot 
\overline{G}_{k(s,a)}(s, a)
\Big)^p
\bigg)^{1/p}\\
=& \norm{\costp \cdot \overline{G}_k + \costr \cdot  (\overline{\chi}^{\targetpi}_{\overline \beta} - \overline F_k)}_p,
\end{align*}
which concludes the proof.

\subsection{Discussion on Choosing $\delta$}
\label{appendix.off.discussion}

\looseness-1While $\epsP$ can be set to small values, making the corresponding constraint in \eqref{prob.off} a relatively weak condition, it is important to note that its value controls parameters of MDP $\widehat M$ that are important for practical considerations in the offline setting. 
Moreover, since $\epsP$ is a parameter in the optimization problems \eqref{prob.on} (and \eqref{prob.on.reformulated}), its value is also important for the online setting. 

In the case of attacks on a learning agent, our results have dependency on the agent's regret or number of suboptimal steps, which in turn depend on the properties of MDP $\widehat M$. For example, if the agent adopts UCRL as its learning procedure, its regret will depend on the diameter of $\widehat M$. Hence, $\epsP$ should be adjusted based on time horizon $T$, so that the parameters of MDP $\widehat M$ relevant for the agent's regret do not outweigh time horizon $T$. 

In the case of attacks on a planning agent with average reward criteria, setting $\epsP$ to small values could result in a solution $\widehat M$ that has a large mixing time, in which case the score $\rho$ might not approximate well the average of obtained rewards in a finite horizon (e.g., see \cite{even2005experts}).
This means that the choice of $\epsP$ should account for the finiteness of time horizon in practice. 

We leave a more detailed analysis that includes these considerations for future work.

\section{Proofs For Online Attacks (Section \ref{sec.onlineattacks})}
\label{appendix.on}
This section contains proof of our results for  Lemma~\ref{lemma.on.subopt.missmatch},
Theorem~\ref{theorem.on.regret},
and Theorem~\ref{theorem.on.subopt}.

\subsection{Proof of Lemma~\ref{lemma.on.subopt.missmatch}}
This lemma is based on the simple observation: when a learner draws its experience from an MDP $M$ that has $\targetpi$ as its $\epsilon$-robust optimal policy, instantiating $\subopt(T, M, \epsilon')$ with $\epsilon' = \epsilon$ will give us the number of times the learner deviates from $\targetpi$. 
In particular, we need to show that $\avgmissm(T) = \frac{1}{T} \subopt(T, M, \epsilon)$ when $\targetpi$ is an $\epsilon$-robust optimal policy. By using the definition of $\subopt(T, M, \epsilon')$ with $\epsilon' = \epsilon$, we obtain:
\begin{align*}
\subopt(T, M, \epsilon) = \sum_{t=0}^{T-1} \ind{a_t \notin \{ \pi(s_t)~|~\rho^\pi \ge \rho^{\pi^*} - \epsilon \} }.
\end{align*}
Since $\targetpi$ is $\epsilon$-robust optimal in $M$, the only $\pi$ satisfying $\rho^\pi \ge \rho^{\pi^*} - \epsilon$ is $\targetpi$ itself. This means that we have
\begin{align*}
\subopt(T, M, \epsilon) &= \sum_{t=0}^{T-1} \ind{a_t \ne \targetpi(s_t) }\\
&= T \cdot \avgmissm(T),
\end{align*}
which proves the claim.


\subsection{Proof of Theorem \ref{theorem.on.regret}: Average Reward Criteria, $\gamma=1$}
We need to prove bounds on the expected cost and average mismatches of the online attack against a regret-minimization learner.

Since $\widehat{M} = (S, A, \widehat{R}, \overline{P})$ is a solution to the optimization problem \eqref{prob.on}, $\targetpi$ is $\epsilon$-robust optimal in $\widehat{M}$. Notice that the learner receives feedback from the MDP $\widehat{M}$. Using Lemma~\ref{lemma.on.known.nontarget} we can obtain the following bound on the expected average mismatches:
$$\expct{\avgmissm(T)} \le
\frac{\widehat{\mu}_{\textnormal{max}}}{\epsilon \cdot T} \cdot \Big(\expct{\regret(T, \widehat{M})} + 2\norm{\widehat{V}^{\targetpi}}_\infty\Big).
$$
Note that $\widehat{V}^{\targetpi} = \overline{V}^{\targetpi}$ and we could also substitute $\widehat{V}^{\targetpi}$ with $\overline{V}^{\targetpi}$ in the bound.

Next, we analyze the expected attack cost. Since $\overline{R}(s, a) = \widehat{R}(s, a), \overline{P}(s, a, s') = \widehat{P}(s, a, s')$ for $s, s', a = \pi_T(s)$, we have
\begin{align*}
    &\expct{\sum_{t=0}^{T-1} \bigg(
\costr \cdot \big|\widehat{R}_t(s_t,a_t) - \overline{R}(s_t,a_t)\big|
+ 
\costp \cdot \sum_{s'}\big|\widehat{P}_t(s_t,a_t,s') - \overline{P}(s_t,a_t,s')\big|
    \bigg)^p}\\
    =& \expct{\sum_{t=0}^{T-1} \ind{a_t \ne \pi(s_t)}
    \bigg(
    \costr \cdot \big|\widehat{R}_t(s_t,a_t) - \overline{R}(s_t,a_t)\big|
+ 
\costp \cdot \sum_{s'}\big|\widehat{P}_t(s_t,a_t,s') - \overline{P}(s_t,a_t,s')\big|
    \bigg)^p}\\
    \le &
    \Big(\cost(\widehat{M}, \overline{M}, \costr, \costp, \infty)
    \Big)^p \cdot T \cdot \expct{\avgmissm(T)}\\
        \le &
    \Big(\cost(\widehat{M}, \overline{M}, \costr, \costp, \infty)
    \Big)^p \cdot
    \frac{\widehat {\mu}_{\textnormal{max}}}{\epsilon } \cdot \Big(\expct{\regret(T, \widehat{M})} + 2\norm{\widehat{V}^{\targetpi}}_\infty\Big).
\end{align*}{} 
Since $p  \ge 1$, the function $f(x) = x^{1/p}$ is concave.  By Jensen's inequality, this means that $\expct{f(X)} \le f(\expct{X})$, where $X$ is a random variable. We can write $\expct{\avgcost(T)}$ to be
\begin{align*}
     &= \frac{1}{T}\expct{\left(\sum_{t=0}^{T-1} 
    \bigg(
\costr \cdot \big|\widehat{R}_t(s_t,a_t) - \overline{R}(s_t,a_t)\big|
+ 
\costp \cdot \sum_{s'}\big|\widehat{P}_t(s_t,a_t,s') - \overline{P}(s_t,a_t,s')\big|
    \bigg)
    ^p\right)^{1/p}}\\
    &\le \frac{1}{T}\expct{\sum_{t=0}^{T-1} 
    \bigg(
\costr \cdot \big|\widehat{R}_t(s_t,a_t) - \overline{R}(s_t,a_t)\big|
+ 
\costp \cdot \sum_{s'}\big|\widehat{P}_t(s_t,a_t,s') - \overline{P}(s_t,a_t,s')\big|
    \bigg)
    ^p}^{1/p}\\
    &\le \frac{\cost(\widehat{M}, \overline{M}, \costr, \costp, \infty)}{T} 
    \cdot \bigg(
\frac{\widehat{\mu}_{\textnormal{max}}}{\epsilon} \cdot \Big(\expct{\regret(T, \widehat{M})} + 2\norm{\widehat{V}^{\targetpi}}_\infty\Big)
    \bigg)^{1/p}.
\end{align*}{}

\subsection{Proof of Theorem \ref{theorem.on.subopt}: Discounted Reward Criteria, $\gamma<1$}
We need to prove bounds on the expected cost and average mismatches of the online attack against a learner with a bounded number of suboptimal steps.

Since $\widehat{M} = (S, A, \widehat{R}, \overline{P})$ is a solution to the optimization problem \eqref{prob.on}, $\targetpi$ is $\epsilon$-robust optimal in $\widehat{M}$. Notice that the learner receives feedback from the MDP $\widehat{M}$. Using Lemma~\ref{lemma.on.subopt.missmatch} we can obtain the following expected average mismatches:
$$\expct{\avgmissm(T)} =
\frac{1}{T} \cdot \expct{\subopt(T, \widehat{M}, \epsilon)}.$$

Next, we analyze the expected attack cost. Since $\overline{R}(s, a) = \widehat{R}(s, a), \overline{P}(s, a, s') = \widehat{P}(s, a, s')$ for $s, s', a = \pi_T(s)$, we have
\begin{align*}
    &\expct{\sum_{t=0}^{T-1} \bigg(
\costr \cdot \big|\widehat{R}_t(s_t,a_t) - \overline{R}(s_t,a_t)\big|
+ 
\costp \cdot \sum_{s'}\big|\widehat{P}_t(s_t,a_t,s') - \overline{P}(s_t,a_t,s')\big|
    \bigg)^p}\\
    =& \expct{\sum_{t=0}^{T-1} \ind{a_t \ne \pi(s_t)}
    \bigg(
    \costr \cdot \big|\widehat{R}_t(s_t,a_t) - \overline{R}(s_t,a_t)\big|
+ 
\costp \cdot \sum_{s'}\big|\widehat{P}_t(s_t,a_t,s') - \overline{P}(s_t,a_t,s')\big|
    \bigg)^p}\\
    \le &
    \Big(\cost(\widehat{M}, \overline{M}, \costr, \costp, \infty)
    \Big)^p \cdot T \cdot \expct{\avgmissm(T)}\\
        \le &
    \Big(\cost(\widehat{M}, \overline{M}, \costr, \costp, \infty)
    \Big)^p \cdot \expct{\subopt(T, \widehat{M}, \epsilon)}.
\end{align*}{}
Since $p  \ge 1$, the function $f(x) = x^{1/p}$ is concave.  By Jensen's inequality, this means that $\expct{f(X)} \le f(\expct{X})$, where $X$ is a random variable. We can write $\expct{\avgcost(T)}$ to be
\begin{align*}
     &= \frac{1}{T}\expct{\left(\sum_{t=0}^{T-1} 
    \bigg(
\costr \cdot \big|\widehat{R}_t(s_t,a_t) - \overline{R}(s_t,a_t)\big|
+ 
\costp \cdot \sum_{s'}\big|\widehat{P}_t(s_t,a_t,s') - \overline{P}(s_t,a_t,s')\big|
    \bigg)
    ^p\right)^{1/p}}\\
    &\le \frac{1}{T}\expct{\sum_{t=0}^{T-1} 
    \bigg(
\costr \cdot \big|\widehat{R}_t(s_t,a_t) - \overline{R}(s_t,a_t)\big|
+ 
\costp \cdot \sum_{s'}\big|\widehat{P}_t(s_t,a_t,s') - \overline{P}(s_t,a_t,s')\big|
    \bigg)
    ^p}^{1/p}\\
    &\le \frac{\cost(\widehat{M}, \overline{M}, \costr, \costp, \infty)}{T} 
    \cdot \Big(\expct{\subopt(T, \widehat{M}, \epsilon)}\Big)^{1/p}.
\end{align*}{}

}
}
{}
\end{document}